\documentclass[]{opendatalab}

\PassOptionsToPackage{numbers, compress}{natbib}

\usepackage[utf8]{inputenc} 
\usepackage[T1]{fontenc}    
\usepackage{hyperref}       
\usepackage{url}            
\usepackage{booktabs}       
\usepackage{amsfonts}       
\usepackage{nicefrac}       
\usepackage{microtype}      
\usepackage{xcolor}         
\usepackage{amsmath}
\usepackage{graphicx}
\usepackage{lineno}

\usepackage{multirow}
\usepackage{multicol}
\usepackage{colortbl}
\usepackage{adjustbox}
\usepackage{caption}
\usepackage{subcaption}
\usepackage{amsmath}
\usepackage{todonotes}
\usepackage{pifont}
\usepackage{array}
\usepackage{ragged2e}
\usepackage{booktabs}
\usepackage{wrapfig}

\usepackage{amsmath}
\usepackage{adjustbox}
\usepackage{multirow}
\usepackage{multicol}
\usepackage{booktabs}
\usepackage{setspace}
\usepackage{hyperref}
\usepackage{threeparttable}
\usepackage{xcolor}
\usepackage{colortbl}
\usepackage{float}
\usepackage{enumitem}
\usepackage{array}
\usepackage{arydshln}
\usepackage{makecell}
\usepackage{amsmath}
\usepackage{tcolorbox}
\usepackage{graphicx}
\usepackage{enumitem}
\usepackage{standalone}
\usepackage{subcaption} 
\usepackage{booktabs}
\usepackage{graphicx}
\usepackage{listings}
\usepackage{xcolor}
\usepackage{tcolorbox}
\usepackage{tabularx}
\tcbuselibrary{skins, breakable}
\definecolor{groupgray}{gray}{0.92}
\newtcolorbox{promptbox}[1]{
    colback=gray!5,            
    colframe=black!75,         
    coltitle=white,            
    title=\textbf{#1},         
    fontupper=\ttfamily\small, 
    breakable,                 
    verbatim,                  
    boxrule=0.5mm,
    arc=2mm,
    top=2mm, bottom=2mm, left=2mm, right=2mm
}

\usepackage{algorithm}
\usepackage{algpseudocode}
\usepackage{colortbl}

\usepackage{amssymb}
\usepackage{mathtools}
\usepackage{amsthm}

\usepackage{colortbl}
\usepackage{nicematrix}

\usepackage{amsthm}
\usepackage{amsmath,amsfonts,bm}
\usepackage{nicefrac}
\usepackage{array}

\usepackage{url}
\usepackage{pbox}

\definecolor{bluex}{rgb}{0.27, 0.42, 0.81}
\definecolor{purplex}{HTML}{9564bf}
\definecolor{red3}{HTML}{C52A20}
\definecolor{red2}{HTML}{B36A6F}
\definecolor{red1}{HTML}{FFb5b5}
\definecolor{purple}{HTML}{B36A6F}
\definecolor{darkyellow}{HTML}{D5BA82}
\definecolor{blue1}{HTML}{508AB2}
\definecolor{blue2}{HTML}{C4E4E3}
\definecolor{green1}{HTML}{A1D0C7}
\definecolor{green2}{HTML}{BFF6BA}
\definecolor{green3}{HTML}{028100}
\definecolor{teal}{HTML}{508AB2}
\definecolor{purple1}{HTML}{8d3a94}

\usepackage{pifont}

\usepackage{tcolorbox}
\tcbuselibrary{listings,theorems}

\theoremstyle{plain}

\theoremstyle{definition}

\theoremstyle{remark}

\setlist[itemize,enumerate,description]{nosep, left=0mm, itemsep=0.5mm}

\tcbset{
  myframe/.style={
    width=\linewidth,
    boxrule=1pt,           
    colframe=black!75,     
    colback=gray!10,       
    arc=2mm,
    left=2mm, right=2mm,
    top=0.2mm, bottom=0.2mm,
    enhanced
  }
}

\newtcolorbox{infobox}[1][]{
  myframe,
  fonttitle=\bfseries,
  title=#1
}

\newtcblisting{codebox}{
  myframe,
  listing only,
  listing options={
    language=Python,
    basicstyle=\ttfamily\scriptsize,
    breaklines=true,
    numbers=none,
    numberstyle=\tiny\color{gray},
    frame=none         
  }
}

\definecolor{mygreen}{HTML}{097969}
\setlist[itemize,enumerate,description]{nosep, left=0mm, itemsep=0.5mm}

\title{Scientific Image Synthesis: Benchmarking, Methodologies, and Downstream Utility}

\author[1,2\dag]{Honglin Lin}
\author[3,2\dag]{Zheng Liu}
\author[4,2\dag]{Chonghan Qin}
\author[2]{Qizhi Pei}
\author[2]{Yu Li}
\author[1,2]{Zhanping Zhong}
\author[1,2]{Xin Gao}
\author[1]{Yanfeng Wang}
\author[2]{Conghui He}
\author[2\ast]{Lijun Wu}

\affiliation[1]{Shanghai Jiao Tong University}
\affiliation[2]{OpenDataLab, Shanghai Artificial Intelligence Laboratory}
\affiliation[3]{Peking University}
\affiliation[4]{The University of Hong Kong}

\abstract{
While synthetic data has proven effective for improving scientific reasoning in the text domain, multimodal reasoning remains constrained by the difficulty of synthesizing scientifically rigorous images. Existing Text-to-Image (T2I) models often produce outputs that are visually plausible yet scientifically incorrect, resulting in a persistent visual–logic divergence that limits their value for downstream reasoning.
Motivated by recent advances in next-generation T2I models, we conduct a systematic study of scientific image synthesis across generation paradigms, evaluation, and downstream use. We examine both direct pixel-based generation and programmatic synthesis, and instantiate ImgCoder as a structured implementation of the code-driven workflow, following an explicit “understand → plan → code” prompting strategy to encourage clearer structural specification.
To rigorously assess scientific correctness, we introduce SciGenBench, which evaluates generated images based on information utility and logical validity. Our evaluation reveals systematic failure modes in pixel-based models and highlights a fundamental expressiveness–precision trade-off. 
Finally, we show that fine-tuning Large Multimodal Models (LMMs) on rigorously verified synthetic scientific images yields consistent reasoning gains, with potential scaling trends analogous to the text domain, validating high-fidelity scientific synthesis as a viable path to unlocking massive multimodal reasoning capabilities.
}

\date{\today}
\metadata[Equal contribution]{Honglin Lin, Zheng Liu, Chonghan Qin}
\correspondence{Lijun Wu, \email{wulijun@pjlab.org.cn}}
\metadata[Project Page]{\url{https://SciGenbench.github.io}}

\begin{document}
\maketitle

\begin{figure*}
    \centering
    \includegraphics[width=0.7\linewidth]{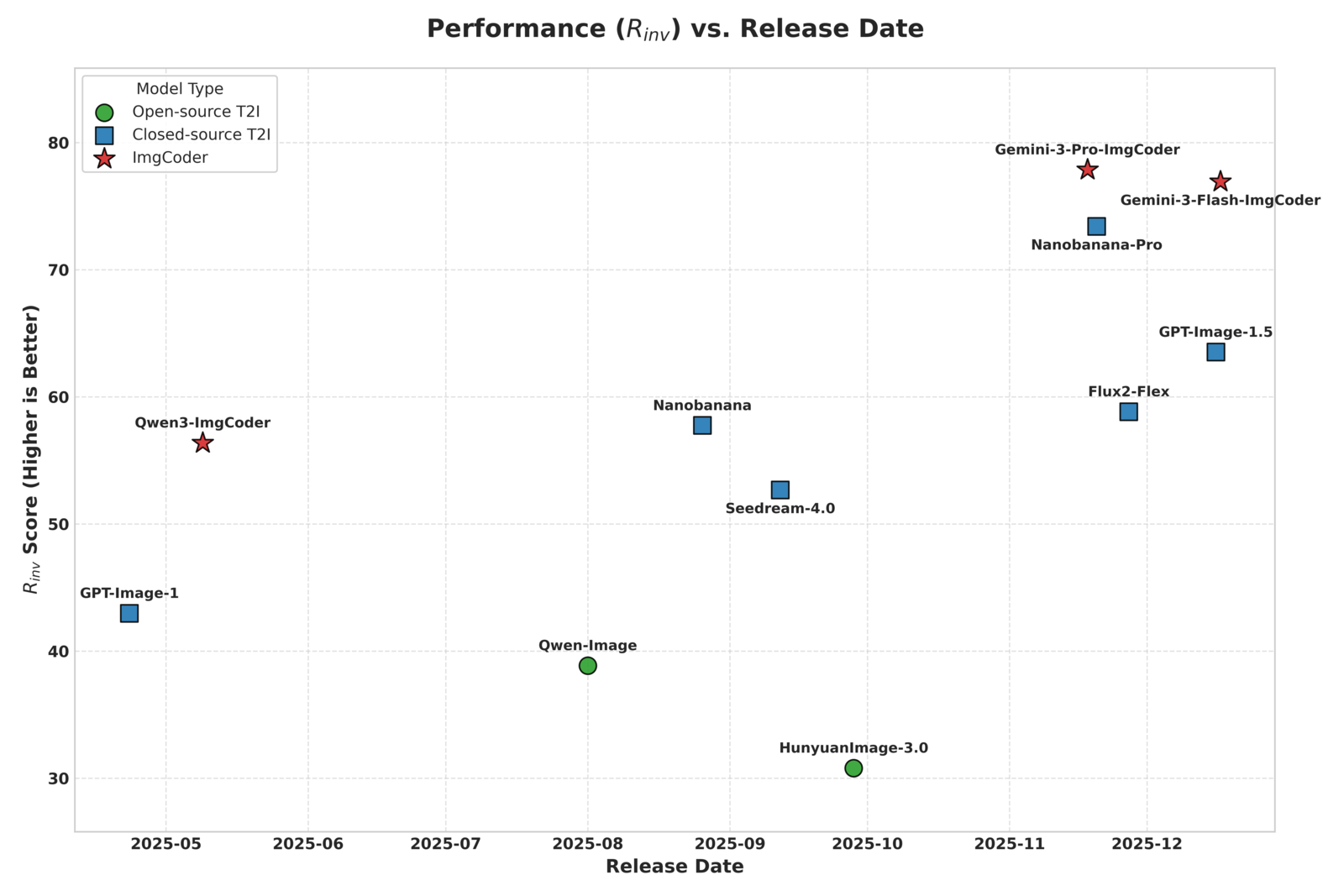}
\caption{\textbf{Performance Evolution on SciGenBench.} Inverse Validation Rate ($R_{\mathrm{inv}}$) over model release dates. ImgCoder (red stars) consistently outperforms contemporary pixel-based baselines.}
\label{fig:method}
\end{figure*}

\section{Introduction}
\label{sec:intro}


With the advancement of Large Multimodal Models (LMMs)~\cite{gemini3,gpt5,qwen3vl}, enabling robust reasoning in scientific domains such as mathematics, physics, and engineering has become an increasingly important goal~\cite{reason_survey,xiang2025seephys,mathvision}. In the text-only setting, large-scale synthetic data has proven effective in alleviating data bottlenecks and improving scientific reasoning~\cite{openthoughts,kpmath,openr1}. However, multimodal scientific reasoning remains significantly underdeveloped. Beyond data scarcity, a fundamental challenge lies in the difficulty of synthesizing scientifically rigorous images. Unlike natural images, scientific visuals must satisfy strict geometric, physical, and relational constraints, which existing Text-to-Image (T2I) models often fail to enforce, producing visually plausible yet scientifically incorrect results.

The recent emergence of next-generation T2I models~\cite{qwenimage,gptimage,seedream}, such as \texttt{Nanobanana-Pro}~\cite{nanobananapro}, has renewed interest in this challenge. These models promise improved semantic understanding and visual control, raising a natural question: \textit{can modern T2I systems overcome the long-standing limitations of scientific image synthesis and unlock scalable multimodal data generation for reasoning?} To answer this question, we propose a systematic investigation with methodological innovation, benchmarking, and downstream reasoning.

From a methodological perspective, scientific image synthesis can be broadly categorized into two paradigms. Pixel-based T2I models generate images end-to-end~\cite{flux1,qwenimage,emu3,januspro,hunyuan} and offer strong visual expressiveness, but often struggle to reliably satisfy strict structural constraints. In contrast, programmatic approaches~\cite{ni2025viscoder,yang2025scaling} generate images through explicit, executable codes, providing stronger structural control. Building on this paradigm, we propose \textbf{ImgCoder}, a logic-driven framework that follows an ``Understand$\rightarrow$Plan$\rightarrow$Code’’ workflow to decouple reasoning from rendering. Our analysis reveals a fundamental \emph{precision–expressiveness} trade-off between these two paradigms.

To rigorously evaluate scientific image generation, we observe that existing benchmarks and metrics—largely focused on pixel similarity or coarse semantic alignment—are insufficient for assessing scientific correctness, where correctness hinges on precise structural and relational validity. We therefore introduce \textbf{SciGenBench}, a specialized benchmark comprising 1.4K problems that are designed to evaluate the information utility and logical validity of generated scientific images across 5 domains and 25 image types. SciGenBench adopts a hybrid evaluation protocol that combines fine-grained LLM-as-Judge scoring with an automated inverse validation pipeline, shifting the focus from visual aesthetics to information utility and logical rigor. Our evaluation uncovers a pervasive visual–logic divergence in current pixel-based models, while code-driven approaches achieve higher structural precision at the cost of reduced expressiveness. We further categorize these failure modes, providing a systematic understanding of where and why different paradigms break down.


Finally, we examine the downstream utility of synthetic scientific images for multimodal reasoning. By grounding image synthesis in scientifically verified textual sources, the resulting visual–text pairs provide reliable supervision for training. Experiments demonstrate that fine-tuning LMMs on such rigorously verified synthetic data yields consistent improvements in scientific reasoning, with evidence of potential scaling trends similar to those observed in text-only settings. These findings suggest that high-fidelity scientific image synthesis offers a viable and scalable pathway toward advancing multimodal scientific reasoning.

\section{Related Work}

\subsection{Scientific Diagram Generation}

\noindent\textbf{Text-to-Image Models.}
Following traditional UNet-based Latent Diffusion Models (LDMs)~\cite{sdxl,sd3}, Diffusion Transformers (DiT)~\cite{dit,sd3,flux1} and Autoregressive Transformers~\cite{emu3,januspro,bagel} have recently emerged to dominate text-to-image generation. While proprietary systems~\cite{nanobananapro,gptimage} and open-source models~\cite{qwenimage,hunyuan} achieve high visual fidelity, generating structured diagrams directly remains challenging~\cite{visualsfactuality,zhang2026sciflow}, revealing a gap between visual realism and structural correctness.

\noindent\textbf{Code-based Methods.}
Early approaches generate visualization code from captions~\cite{belouadi2023automatikz}, while later works scale instruction-based plotting via curated code corpora~\cite{ni2025viscoder}. Some methods rely on manually defined templates~\cite{yang2025scaling,yang2025effective}, at the cost of limited output diversity.

\subsection{Benchmarking Text-to-Image Generation}

\noindent\textbf{From Fidelity to Reasoning.}
Early benchmarks~\cite{yu2022scaling,tifa,t2icompbench,mjbench,prism-bench} focus on visual fidelity and semantic alignment via automated metrics (e.g., FID, CLIP), while recent works shift toward \textit{Reasoning-Informed} evaluation~\cite{r2ibench}, assessing physical plausibility or contextual consistency in natural scenes (e.g., T2I-ReasonBench~\cite{t2ireasonbench}, RISEBench~\cite{risebench}).

\noindent\textbf{The Gap: Reasoning-Oriented Synthesis.}
In contrast, scientific image synthesis requires explicitly materializing abstract axioms into precise visual structures (e.g., circuit topologies). Unlike natural scenes with implied logic, these tasks demand active computation under strict domain laws.

\begin{figure*}
    \centering
    \includegraphics[width=0.99\linewidth]{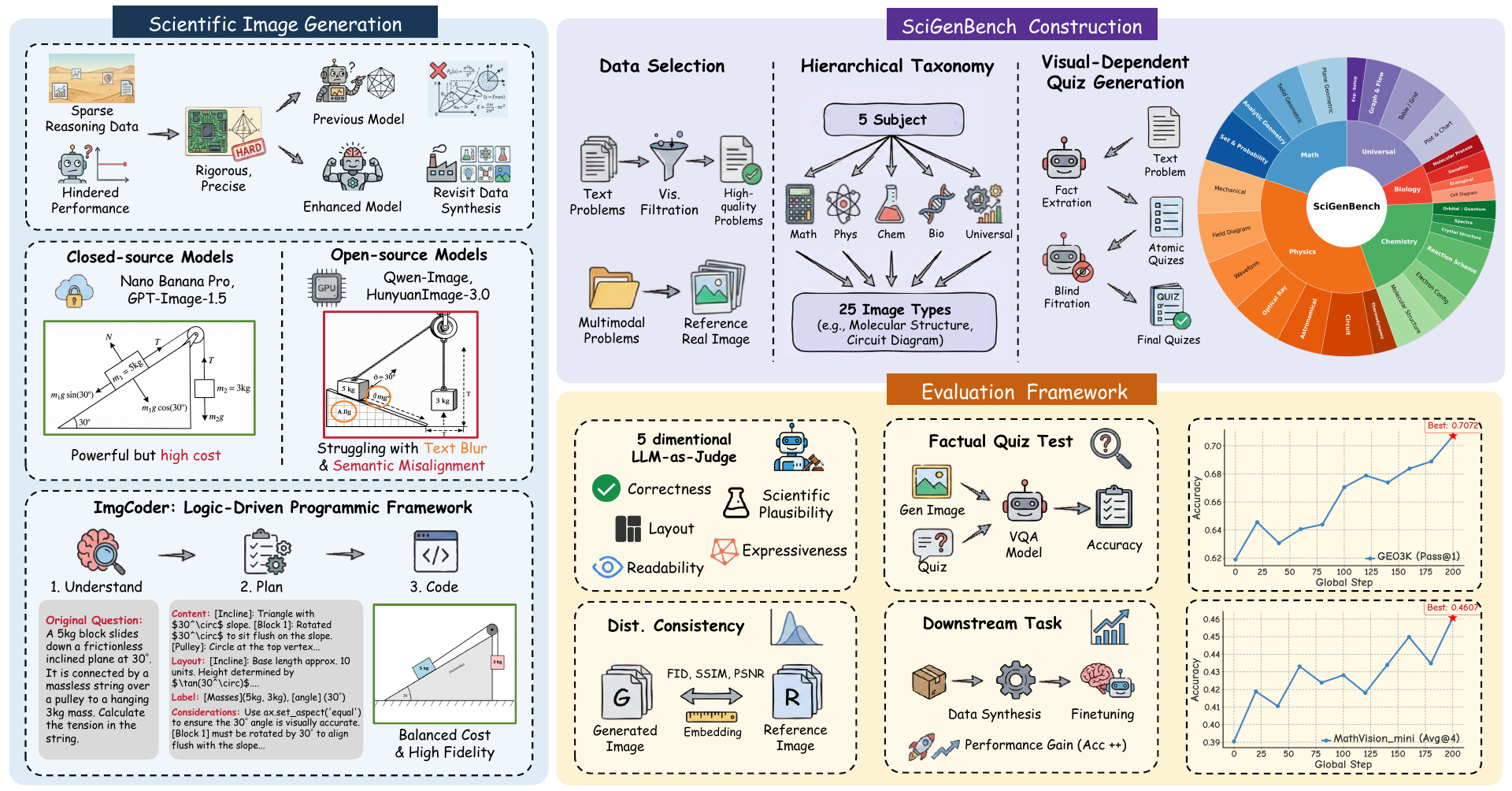}
\caption{\textbf{Methodological Overview.} 
The framework consists of three core components: 
(1) \textbf{Scientific Image Generation} (Left), where we propose ImgCoder, a programmatic approach decoupling planning from implementation to outperform pixel-based baselines; 
(2) \textbf{SciGenBench Construction} (Top Right), a rigorously curated benchmark with a fine-grained taxonomy and atomic quizzes; and 
(3) \textbf{Evaluation Framework} (Bottom Right), a multi-faceted assessment system combining LMM judges, inverse validation, standard metrics, and downstream performance.}
    \label{fig:method}
\end{figure*}

\section{Scientific Image Generation}
\label{sec:frameworks}


We investigate the scientific image synthesis by comparing two different paradigms. As shown in Figure~\ref{fig:method} (Left), pixel-based models synthesize images end-to-end from text, while code-driven approaches generate executable specifications that are deterministically rendered. Our method, \textbf{ImgCoder}, adopts the code-driven paradigm and implements a logic-first \emph{``Understand $\rightarrow$ Plan $\rightarrow$ Code’’} workflow, enabling explicit planning and structured execution for science-intensive visuals.

\subsection{Task Definition}
\label{subsec:task_def}


Let $\mathcal{T}$ denote the space of scientific descriptions and $\mathcal{I}$ the image space. We formulate scientific image generation as a \textit{conditional generation} problem with a parameterized model $\mathcal{M}_\theta: \mathcal{T} \rightarrow \mathcal{I}$. Unlike open-domain synthesis, valid images are constrained by a set of latent scientific axioms $\mathcal{A}$.
Given a query $T \in \mathcal{T}$, the goal is to generate an image $I^*$ that maximizes the likelihood of the reasoning outcome $y$ under a downstream solver $\mathcal{S}$:
\setlength{\abovedisplayskip}{3pt}
\setlength{\belowdisplayskip}{3pt}
\begin{equation}
    I^* = \operatorname*{arg\,max}_{I \sim \mathcal{M}_\theta(\cdot \mid T)} 
    P_{\mathcal{S}}(y \mid I, T, \mathcal{A}),
\end{equation}
where $y$ is determined by $\mathcal{A}$ and $P_{\mathcal{S}}$ denotes the probability assigned by the solver to $y$. This objective encourages images that are not only visually coherent but also structurally faithful for downstream reasoning.

\subsection{Pixel-based Synthesis Framework}
\label{subsec:pixel_framework}

This framework represents the direct synthesis paradigm, treating scientific image generation as an end-to-end translation task from textual descriptions to visual pixel space.
We leverage SOTA T2I models as generators, including both proprietary systems (e.g., \texttt{Nanobanana-Pro}, \texttt{GPT-Image-1.5}, \texttt{Seedream-4.0}, \texttt{Flux2}) known for their robust instruction following, and leading open-source models (e.g., \texttt{Qwen-Image}, \texttt{HunyuanImage-3.0}) representing accessible multimodal capabilities.

To bridge the gap between general T2I capabilities and STEM requirements, we employ a constraint-injection strategy (Prompt~\ref{pmt:t2igen}) for image generation. This pipeline regulates generation by enforcing \emph{information fidelity} (explicit visualization of all entities), applying strict \emph{negative constraints} (preventing solution leakage like calculation steps), and ensuring \emph{stylistic standardization} (prioritizing a clean, textbook-style aesthetic).

\subsection{ImgCoder: Code-based Synthesis Framework}
\label{subsec:imgcoder}
While pixel-based models provide flexible end-to-end synthesis, they are inherently constrained by the underlying diffusion paradigm, which makes it difficult to explicitly enforce structured constraints and logical consistency during generation. Code-driven approaches, in contrast, decouple structural specification from visual rendering, allowing deterministic control over geometry and semantics. We therefore explore a programmatic paradigm and refer to the large-model-based code generation pipeline for deterministic image rendering as \textbf{ImgCoder}. Rather than proposing a new generation paradigm, our goal is to systematically instantiate and evaluate the code-driven workflow under a unified experimental protocol. Concretely, we implement an \emph{``Understand $\rightarrow$ Plan $\rightarrow$ Code’’} baseline, where the model generates executable code (e.g., Python) for structured image rendering. To reduce layout inconsistencies commonly observed in direct text-to-code generation, ImgCoder incorporates an explicit reasoning stage prior to synthesis, encouraging the model to outline the visualization intent before emitting code. Concretely, the planning phase explicitly defines four aspects of the target figure: \textbf{(1) Image Content}, by exhaustively identifying all geometric entities, physical components, and their logical relationships; \textbf{(2) Layout}, by pre-planning coordinate systems and topological arrangements to prevent visual clutter and unintended overlaps; \textbf{(3) Labels}, by determining both the semantic content and precise anchor points of textual annotations; and \textbf{(4) Drawing Constraints}, by validating the plan against domain-specific axioms (e.g., geometric rules or physical laws) while strictly avoiding answer leakage.To examine the effectiveness of ImgCoder across model backbones, we implement two variants: \texttt{Qwen3-ImgCoder} and \texttt{Gemini3-ImgCoder}, built upon \texttt{Qwen3-235B-Instruct}~\cite{qwen3} and \texttt{Gemini3}, respectively.

\section{SciGenBench: Benchmarking Scientific Image Synthesis}
\label{sec:benchmark}

We next turn to evaluation. Standard image-generation metrics are not designed for scientific imagery, where even minor structural errors can invalidate the underlying semantics. We therefore introduce \textbf{SciGenBench} to benchmark the paradigms in Section~\ref{sec:frameworks} by measuring the \emph{information utility} and \emph{logical correctness} of generated images. Figure~\ref{fig:method} (Right) overviews our pipeline, which integrates curated data construction, a hierarchical taxonomy, and a hybrid evaluation framework.

\begin{table*}[h!]
    \centering
    \footnotesize
    \caption{\textbf{Evaluation Dimensions for LMM-as-Judge.} Gemini-3-Flash scores images (0-2) across five criteria.}
    \label{tab:metrics}
    \resizebox{\linewidth}{!}{%
    \renewcommand{\arraystretch}{0.93}
    \begin{tabularx}{\linewidth}{l X}
        \toprule
        \textbf{Dimension} & \textbf{Evaluation Criteria \& Focus} \\
        \midrule
        \textbf{Correctness \& Fidelity} & Strict prompt adherence; detects compositional errors (quantity/attribute mismatch). \\
        \addlinespace[0.3em]
        \textbf{Layout \& Precision} & Precision of geometric construction, topological accuracy, and coordinate system alignment. \\
        \addlinespace[0.3em]
        \textbf{Readability \& Occlusion} & Clarity of textual labels, checking for occlusion, garbled text, or background interference. \\
        \addlinespace[0.3em]
        \textbf{Scientific Plausibility} & Conformity to domain-specific axioms (e.g., valency balance, Newton's laws). \\
        \addlinespace[0.3em]
        \textbf{Expressiveness \& Richness} & Ensures holistic context and scenario completeness, avoiding isolated elements. \\
        \bottomrule
    \end{tabularx}
    }
\end{table*}

\subsection{Data Acquisition \& Selection}
SciGenBench is designed with two complementary data sources that serve distinct evaluation purposes: a primary instruction-driven benchmark for assessing synthesized scientific images, and a real-world visual reference set for distributional comparison.

\begin{itemize}
    \item \textbf{Verified Scientific Instructions.}
    The core of SciGenBench is constructed from high-quality scientific text corpora, including \texttt{MegaScience}~\cite{fan2025megascience} and \texttt{WebInstruct-verified}~\cite{webinstruct_verified}, which are selected for their logical correctness and factual rigor. To ensure suitability for visual generation, we apply a visualizability filtration (Prompt~\ref{pmt:curation}) to remove non-visual content (e.g., abstract derivations), retaining only descriptions with concrete, imageable structures.
    
    \item \textbf{Real-World Visual Reference.}
    To contextualize synthetic scientific images against authentic visual data, we incorporate a vision-optional subset of \texttt{SeePhys}~\cite{xiang2025seephys} as a human-authored reference set, denoted \textbf{SciGenBench-SeePhys}. This subset participates in all evaluation dimensions and serves as the exclusive ground truth for reference-based standard image metrics used in real--synthetic comparison.
\end{itemize}


    

\subsection{Hierarchical Taxonomy}
To establish a structured data distribution, we organize SciGenBench using a rigorous two-level \textit{``Subject–Image Type’’} taxonomy. We employ \texttt{Gemini-3-Flash} as a unified data curator (Prompt~\ref{pmt:curation}) to perform visualizability filtration and fine-grained classification jointly within a single inference pass, enabling scalable and consistent annotation. 
The resulting taxonomy is defined along two hierarchical dimensions. \textbf{(1) 5 Subjects} comprise Mathematics, Physics, Chemistry, Biology categories, and a \textit{Universal} category that captures cross-domain visual structures. \textbf{(2) 25 Image Types} include fine-grained categories reflecting domain-specific visual conventions, such as \textit{molecular structure} in Chemistry or \textit{circuit diagram} in Physics, as well as generic types under the Universal subject, including \textit{chart \& graph} and \textit{experimental setup}. Detailed definitions of all image types are provided in Appendix~\ref{app:taxonomy}.
\subsection{Visual-Dependent Quiz Generation}
\label{sec:quizgen}
To ensure that generated images provide high information utility and are strictly indispensable for problem solving, we design an automated \emph{``Generate–Filter–Select’’} pipeline to construct atomic, visually grounded quizzes.

\noindent \textbf{Fact-based Question Formulation.}
Using \texttt{Gemini-3-Flash} with Prompt~\ref{pmt:quizgen}, we first analyze each source instruction to extract verifiable factual elements, such as numerical values, geometric relations, or domain-specific properties (e.g., electron configurations). These elements are then converted into atomic questions, each targeting a single, concrete piece of visual information.

\noindent \textbf{Blind Filtration for Visual Necessity.}
To eliminate pseudo-multimodal questions that can be solved without visual input, we introduce a \textit{blind filtration} mechanism. Specifically, \texttt{GPT-5-nano} is employed as a blind solver that receives only the above question text, without access to image. Each question is evaluated across 4 independent trials; those answered correctly in all trials are discarded, as they indicate text leakage or reliance on commonsense rather than visual evidence.

\noindent \textbf{Density-based Selection and Human Review.}
To prioritize information-rich samples, we retain images associated with the highest number of valid atomic quizzes within each image type. The resulting quiz set is further reviewed by expert annotators to ensure logical consistency, scientific rigor, and the absence of hallucinations.
    
    

\subsection{Evaluation Framework}
\label{sec:eval_framework}

Evaluating scientific image generation poses unique challenges, as traditional pixel-level metrics fail to capture logical correctness and scientific factuality. To establish a rigorous standard, we adopt a hybrid evaluation strategy comprising automated judge, inverse validation, traditional metrics and downstream utility assessment, as shown in Figure~\ref{fig:method} (Bottom Right).

\noindent\textbf{Multi-dimensional LMM-as-Judge.} 
Following the structured rubric defined in Prompt~\ref{pmt:llm_judge}, we employ \texttt{Gemini-3-Flash} as an automated evaluator. For each generated image, the judge produces a reasoning critique along with a score $s \in \{0,1,2\}$ for each evaluation dimension specified in Table~\ref{tab:metrics}, enabling fine-grained and holistic assessment of visual logical correctness.

\noindent\textbf{Inverse Quiz Validation.}
To quantify whether a generated image faithfully encodes its intended information, we introduce an inverse validation metric. Let $\mathcal{Q}_I$ denote the set of atomic quizzes associated with image $I$, and $\mathcal{V}(I, q) \in \{0,1\}$ indicate the correctness of a strong VQA model’s response to question $q \in \mathcal{Q}_I$. We define the \textbf{inverse validation rate} ($\mathcal{R}_{\text{inv}}$) as the proportion of images for which \emph{all} associated quizzes are answered correctly:
\begin{equation}
\small
\mathcal{R}_{\text{inv}} = \frac{1}{|\mathcal{D}|} \sum_{I \in \mathcal{D}}
\mathbb{I}\left( \sum_{q \in \mathcal{Q}_I} \mathcal{V}(I, q) = |\mathcal{Q}_I| \right),
\end{equation}
where $\mathbb{I}(\cdot)$ is the indicator function and $\mathcal{D}$ denotes the evaluation set. To ensure that validation failures primarily reflect informational deficiencies in the image rather than limitations of the solver, we use \texttt{Gemini-3-Flash} as the VQA engine.

\noindent\textbf{Reference-based Standard Metrics.}
For comparability with prior T2I research, we also report conventional automated metrics (FID, PSNR, SSIM) computed against ground-truth images. These metrics quantify the visual domain gap between synthetic and authentic scientific images, but serve as auxiliary references due to the sparse pixel distribution of scientific diagrams. These metrics are applied only to images from SciGenBench-SeePhys.


\noindent\textbf{Data Utility for Training.}
As the ultimate measure of quality, we evaluate the performance gains of LMMs fine-tuned on synthetic images produced by different generation paradigms. This directly assesses the functional utility and logical consistency of synthesized data from a downstream reasoning perspective (see Section~\ref{subsec:downstream}). To prevent textual shortcuts during training, we apply a \emph{multimodal adaptation} strategy (Prompt~\ref{pmt:multimodal_adaptation}), which masks textual cues and enforces reliance on visual evidence.
\begin{table*}[t]
    \centering
    \caption{\textbf{Overall results on SciGenBench.}
    We report the inverse validation rate ($\mathcal{R}_{\text{inv}}$) and the LMM-as-Judge scores.
    Standard-metrics are computed on the real-image SeePhys subset.}
    \label{tab:main_results}
    \resizebox{0.95\linewidth}{!}{%
        \renewcommand{\arraystretch}{0.92}
    \begin{tabular}{l c ccccc cccc}
        \toprule
        \multirow{2}{*}{\textbf{Model}} &
        \multirow{2}{*}{\textbf{$\mathcal{R}_{\text{inv}}$} (\%) $\uparrow$} &
        \multicolumn{5}{c}{\textbf{LMM-as-Judge (0--2)} $\uparrow$} &
        \multicolumn{4}{c}{\textbf{Standard-metrics}} \\

        \cmidrule(lr){3-7}
        \cmidrule(lr){8-11}

        & &
        \textbf{C\&F} &
        \textbf{L\&P} &
        \textbf{R\&O} &
        \textbf{SP} &
        \textbf{E\&R} &
        \textbf{PSNR} $\uparrow$ &
        \textbf{SSIM} $\uparrow$ &
        \textbf{CLIP} $\uparrow$ &
        \textbf{FID} $\downarrow$ \\

        \midrule
        \multicolumn{11}{l}{\textbf{Open-source T2I Models}} \\
        \midrule

        HunyuanImage-3.0 &
        30.79 &
        0.39 & 0.78 & 1.44 & 0.56 & 0.81 &
        12.21 & 0.82 & 25.01 & 93.27 \\

        Qwen-Image &
        38.86 &
        0.24 & 0.70 & 1.48 & 0.30 & 0.76 &
        9.63 & 0.78 & 25.02 & 120.42 \\
        
        \midrule
        \multicolumn{11}{l}{\textbf{Closed-source T2I Models}} \\
        \midrule

        GPT-Image-1 &
        42.97 &
        0.57 & 1.37 & 1.90 & 0.84 & 1.19 &
        13.07 & 0.84 & 25.14 & \textbf{77.31} \\

        Seedream-4.0 &
        52.67 &
        0.44 & 0.94 & 1.67 & 0.55 & 0.95 &
        10.65 & 0.74 & 25.02 & 98.22 \\

        Nanobanana &
        57.75 &
        0.43 & 0.92 & 1.60 & 0.60 & 1.15 &
        14.12 & 0.85 & 25.13 & 104.70 \\

        Flux2-flex &
        58.83 &
        0.48 & 1.06 & 1.70 & 0.67 & 1.20 &
        14.11 & 0.85 & 25.10 & 96.74 \\

        GPT-Image-1.5 &
        63.52 &
        0.98 & 1.70 & \underline{1.97} & 1.17 & 1.62 &
        \textbf{14.79} & \textbf{0.88} & 25.16 & 112.52 \\

        Nanobanana-Pro &
        73.41 &
        1.59 & 1.87 & \textbf{1.98} & 1.72 & \textbf{1.93} &
        12.02 & 0.81 & 25.01 & \underline{87.72} \\

        \midrule
        \multicolumn{11}{l}{\textbf{ImgCoder}} \\
        \midrule

        Qwen3-ImgCoder &
        56.38 &
        1.21 & 1.30 & 1.62 & 1.39 & 1.29 &
        \underline{14.71} & \underline{0.86} & \textbf{25.21} & 121.55 \\

        Gemini-3-Flash-ImgCoder &
        \underline{76.93} &
        \underline{1.80} & \underline{1.88} & 1.88 & \underline{1.92} & \underline{1.91} &
        14.63 & 0.85 & \underline{25.18} & 117.83 \\

        Gemini-3-Pro-ImgCoder &
        \textbf{77.87} &
        \textbf{1.82} & \textbf{1.93} & 1.91 & \textbf{1.93} & 1.90 &
        14.59 & \underline{0.86} & 25.16 & 107.67 \\

        \bottomrule
    \end{tabular}%
    }
\end{table*}
\section{Benchmarking Generative Capabilities}

\label{sec:t2i_experiments}
In this section, we aim to address pivotal research questions (RQs) regarding scientific image synthesis through comprehensive benchmarking.
\textbf{RQ1- Generative Capability:} How do current SOTA models perform in scientific image generation task?
\textbf{RQ2 - Paradigm Comparison:} What are the respective trade-offs between generative models and programmatic frameworks?
Detailed experimental setup are provided in Appendix~\ref{apx:setup}.

\subsection{Quantitative Results}
Table~\ref{tab:main_results} presents the comprehensive benchmarking results across our proposed taxonomy. The quantitative analysis reveals several key insights:

\noindent\textbf{Closed- vs. Open-source Performance Gap.}
Closed pixel-based models consistently outperform open-source T2I systems across both inverse validation rate and LMM-as-Judge evaluations, benefiting from large-scale proprietary data and optimized training pipelines. For instance, \texttt{Nanobanana-Pro} achieves a $\mathcal{R}_{\text{inv}}$ of 73.41\% with strong Judge scores on C\&F (1.59) and L\&P (1.87), whereas open-source models such as \texttt{HunyuanImage-3.0} and \texttt{Qwen-Image} remain below 40\% $\mathcal{R}_{\text{inv}}$ and score under 0.8 on these structure-sensitive dimensions. This persistent gap indicates that scaling model size alone is insufficient for scientific diagram generation without stronger inductive biases or explicit constraint mechanisms.


\begin{figure*}[h]
    \centering
    \includegraphics[width=0.9\linewidth]{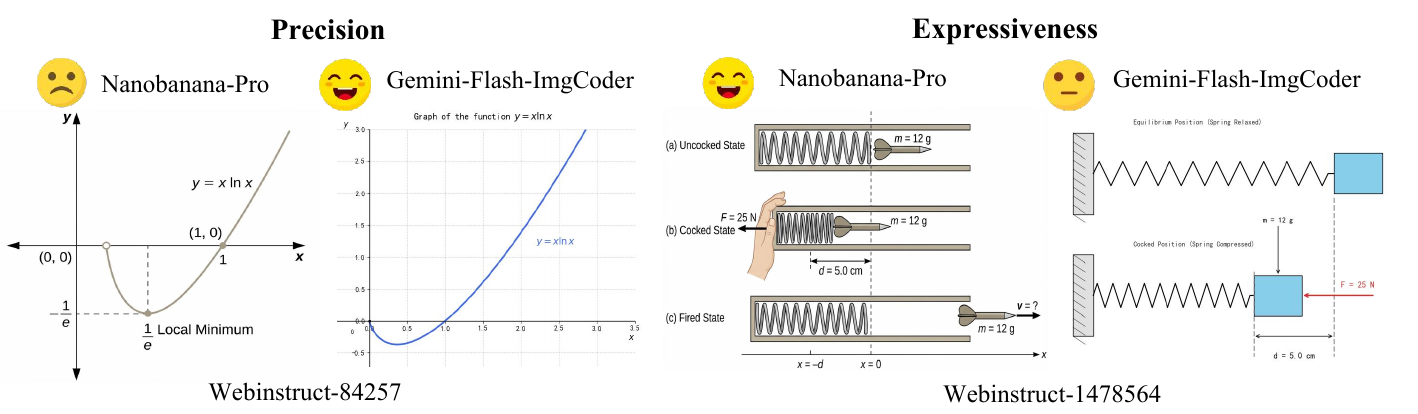}
    \caption{\textbf{Precision vs. Expressiveness Trade-off.} 
  \textit{Left (a):} When plotting the function $y=x \ln x$, pixel-based models produce visually smooth but mathematically inaccurate plots, while code-based methods ensure exactness via execution.
  \textit{Right (b):} Conversely, for physical scenarios like a spring system, pixel-based models offer richer visual expressiveness, whereas code-based outputs remain schematic.}
    \label{fig:precision_expressiveness}
\end{figure*}

\begin{figure*}[h]
    \centering
    \includegraphics[width=0.98\linewidth]{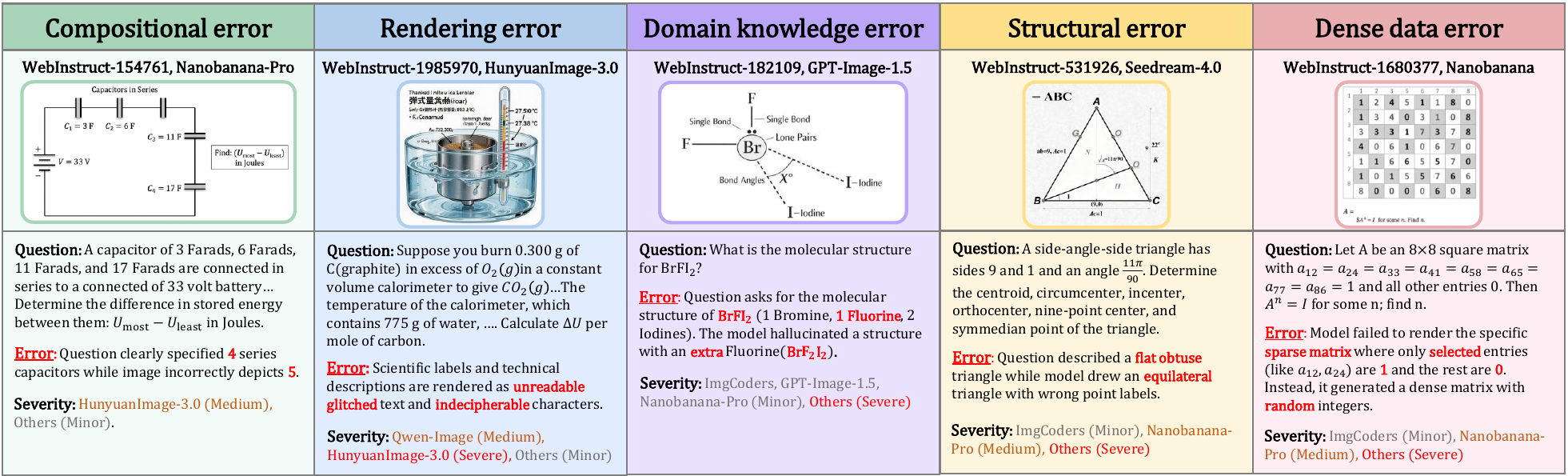}    
    \caption{\textbf{Qualitative Error Taxonomy.} We categorize failures into five modes ranging from low-level visual artifacts to high-level semantic hallucinations. The specific errors are annotated in red.}
    \label{fig:error_cases}
\end{figure*}

\noindent\textbf{Advantages of ImgCoder.}
Interestingly, ImgCoder models consistently achieve the strong inverse validation rates and LMM-as-Judge performance, particularly on structure- and reasoning-sensitive dimensions such as C\&F, L\&P, and SP. Notably, \texttt{Gemini-3-Pro-ImgCoder} attains the top $\mathcal{R}_{\text{inv}}$ (77.87\%), surpassing all pixel-based models. 
Besides, \texttt{Qwen3-ImgCoder} achieves a large improvement over \texttt{Qwen-Image} (56.38\% vs. 38.86\% $\mathcal{R}_{\text{inv}}$) despite being fully open-source, confirming that code-based execution—rather than closed data—drives the observed gains.

\noindent\textbf{Disparity between Evaluation Metrics.}
We observe a clear divergence between perceptual standard metrics and reasoning-oriented evaluations. Models with competitive PSNR or FID scores do not necessarily achieve high inverse validation rates or judge scores. A closer inspection of judge dimensions reveals that perceptual fidelity primarily correlates with E\&R, while exhibiting weak alignment with C\&F and L\&P, which require strict logical consistency and structural correctness. This discrepancy underscores the limitation of previous pixel-level metrics in evaluating scientific diagrams, where visual similarity can mask subtle yet critical factual/relational errors.

\noindent\textbf{Spiral Co-evolution Hypothesis.}
We hypothesize a complementary co-evolution between code-based and pixel-based paradigms. On one hand, code-based methods enforce structural and logical validity, and as LMMs improve, such structured reasoning can transfer to pixel-based T2I models through shared backbones or by using code-rendered synthetic images as training data. Empirically, \texttt{Nanobanana-Pro} (\texttt{Gemini-3-Pro-Image}) and \texttt{Gemini-3-ImgCoder}, which share the \texttt{Gemini-3} backbone, exhibit highly similar diagram construction strategies across many samples (Figure~\ref{fig:precision_expressiveness}, Appendix~\ref{apx:cases}), supporting this transfer. On the other hand, pixel-based models contribute visually diverse and expressive imagery that enriches multimodal training data, which in turn benefits code-based reasoning and LMMs learning, forming a mutually reinforcing spiral between reasoning and synthesis.

\subsection{Qualitative Analysis}

\noindent\textbf{Error Taxonomy.}
As illustrated in Figure~\ref{fig:error_cases}, we categorize observed failures into five primary classes based on a systematic audit.
\textbf{(1) Compositional error:} Misalignment among multiple visual elements, including attribute leakage (e.g., incorrect label binding), spatial relation confusion, or incorrect object counts.
\textbf{(2) Rendering error:} Low-level fidelity issues such as illegible text glyphs, blurred lines, or visually corrupted labels.
\textbf{(3) Structural error:} Violations of geometric logic or topological integrity, including distorted shapes, non-closed curves, or logically inconsistent intersections (e.g., broken parallelism).
\textbf{(4) Dense data error:} Failures in high-information-density scenarios (e.g., tables or matrices), manifested as coordinate drift, axis misalignment, or collapsed rows and columns.
\textbf{(5) Domain knowledge error:} Scientific hallucinations where images appear visually plausible but violate domain-specific laws (e.g., incorrect electron configurations or optical paths).

We observe a hierarchy in failure modes. Traditional text-to-image errors (compositional, rendering) are largely resolved in SOTA models, persisting only in weaker baselines. Domain knowledge error serves as a watershed: while open-source models struggle, top-tier closed-source models have significantly reduced scientific hallucinations. However, structural and dense data errors remain the most persistent bottlenecks. These tasks demand high-precision rendering and information density that challenge the probabilistic nature of diffusion models, causing even \texttt{Nanobanana-Pro} to falter. Notably, \texttt{ImgCoder} outperforms pixel-based SOTAs in these high-precision regimes due to its rigorous code-driven grounding.

\noindent\textbf{Precision-Expressiveness Trade-off.} In Figure~\ref{fig:precision_expressiveness}, the two paradigms exhibit distinct trade-offs. 
While code-based methods may fall slightly short in visual expressiveness—often producing schematic or ``flat" diagrams compared to the rich renderings of \texttt{Nanobanana-Pro} (e.g., the textured spring and hand illustration)—they offer an intrinsic advantage in precision. 
Pixel-based models, constrained by their probabilistic generation nature, fundamentally struggle to guarantee strict mathematical accuracy. 
This is evident in the function plotting task ($y = x \ln x$), where \texttt{Nanobanana-Pro} produces a plausible-looking curve but fails to capture the correct intercepts and extrema.
In contrast, code-based methods leverage deterministic execution engines to ensure the rigorous alignment of geometric shapes and data coordinates, effectively solving the "hallucination" problem in quantitative visualization.



\begin{figure}[t!]
    \centering
    \includegraphics[width=0.9\linewidth]{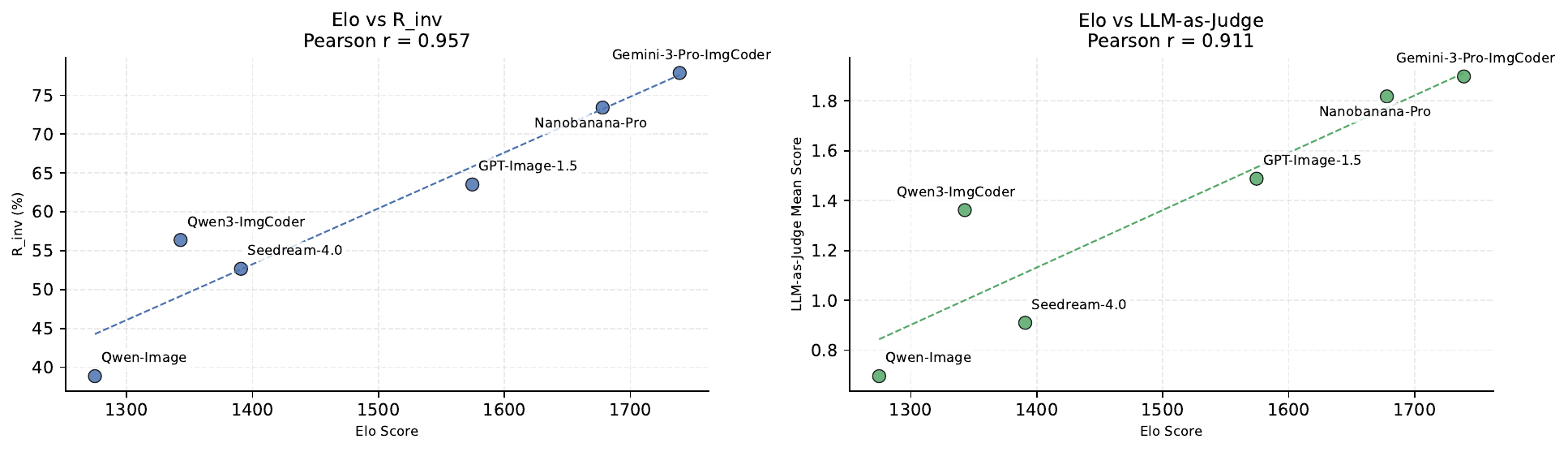}
    \caption{\textbf{Correlation between human-derived Elo scores and automatic evaluation metrics ($\mathcal{R}_{\text{inv}}$ and LMM-as-Judge).} Both metrics show strong alignment with human preferences in terms of absolute scores and relative rankings.}
    \label{fig:human_eval}
\end{figure}
\subsection{Human–Judge Consistency Analysis}

Since automated metrics rely on model capability, we conducted a human evaluation study to assess the reliability of LMM-as-Judge. Ten annotators with STEM backgrounds performed pairwise preference comparisons on the outputs generated by different methods under the same evaluation criteria, the evaluation covered 500 different samples. To balance evaluation coverage and annotation cost, we selected a representative subset of models for comparison. Annotators were asked to choose a winner, or select one of the following options: ``both good,'' ``both bad,'' or ``problematic question.'' Questions marked as problematic were filtered out from the final dataset (\cref{sec:quizgen}).

We then computed the Elo scores of these models ($\sigma = 400$), with initial scores set to 1500 and $K = 32$. Since Elo computation is order-dependent, we shuffled the win–loss pairs and repeated the calculation 100 times, and reported the average scores.
Finally, we conducted a correlation analysis between the Elo scores derived from human preferences and the $\mathcal{R}_{\text{inv}}$ and LMM-as-Judge scores. As shown in~\cref{fig:human_eval}, the scores and rankings of the two metrics are highly consistent with human preferences. Under different evaluation settings, the relative ranking among methods remains stable. These results indicate that, for scientific diagram generation tasks, these metrics can serve as a reasonable proxy for human comparative evaluation, and the selected \texttt{Gemini-3-Flash} model demonstrates reliable discriminative capability.

\subsection{Distributional Gap with Real Images}
\label{subsec:dist_gap}
\noindent \textbf{FID Discrepancy.}
As shown in Table~\ref{tab:main_results}, even top-tier models that avoid overt factual errors still differ subtly from real-image distributions. Models such as \texttt{GPT-Image-1} and \texttt{Nanobanana-Pro} achieve comparatively low FID scores (70.73 and 79.89), due to a concise ``textbook-style'' visual appearance (see Appendix~\ref{apx:cases}).

\begin{figure*}[h] 
    \centering
    \begin{subfigure}[b]{0.32\linewidth}
        \centering
        \includegraphics[width=1.0\linewidth]{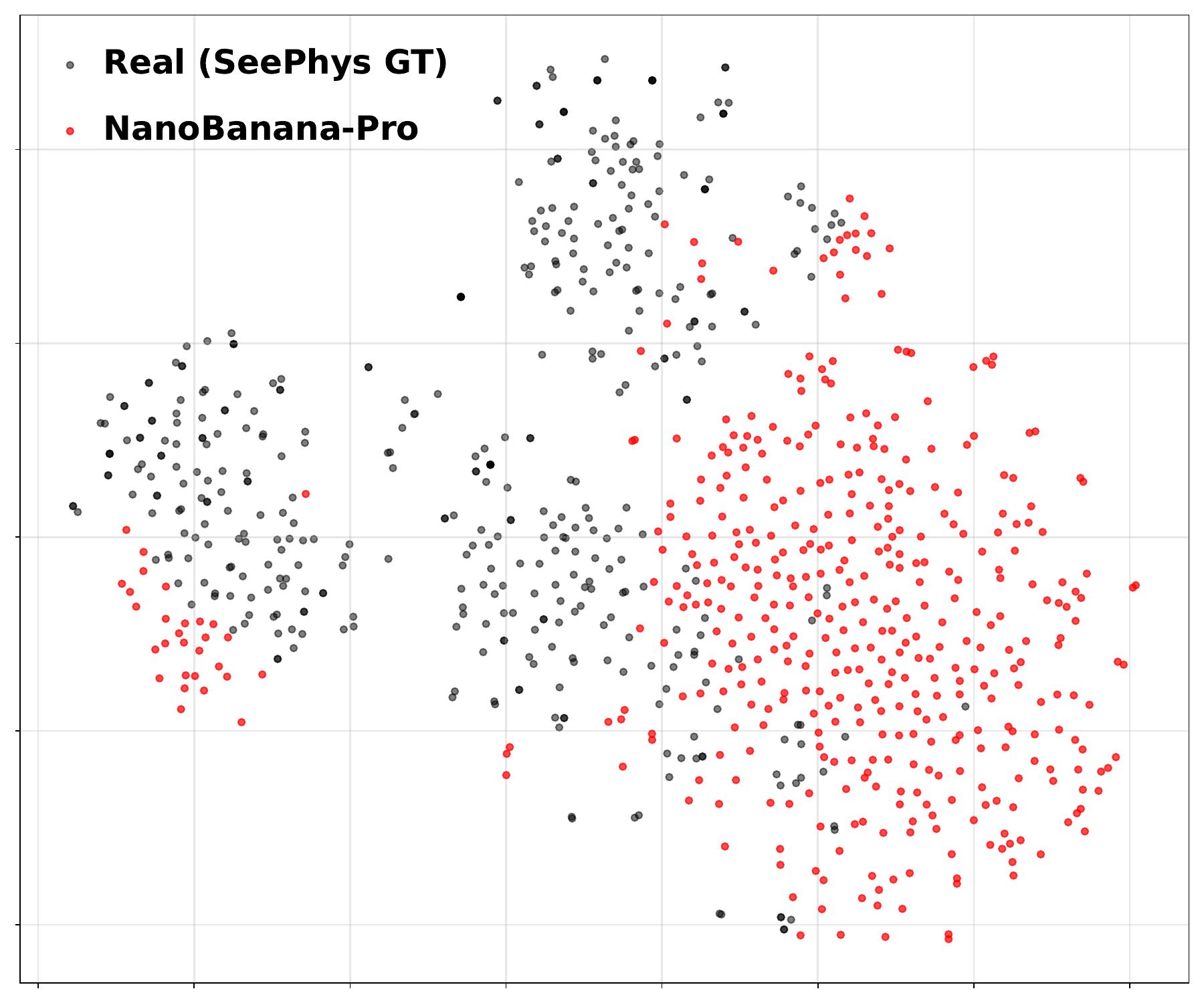}
        \phantomsubcaption
        \label{fig:embedding_nano}
    \end{subfigure}
    \hfill 
    \begin{subfigure}[b]{0.45\linewidth}
        \centering
        \includegraphics[width=1.0\linewidth]{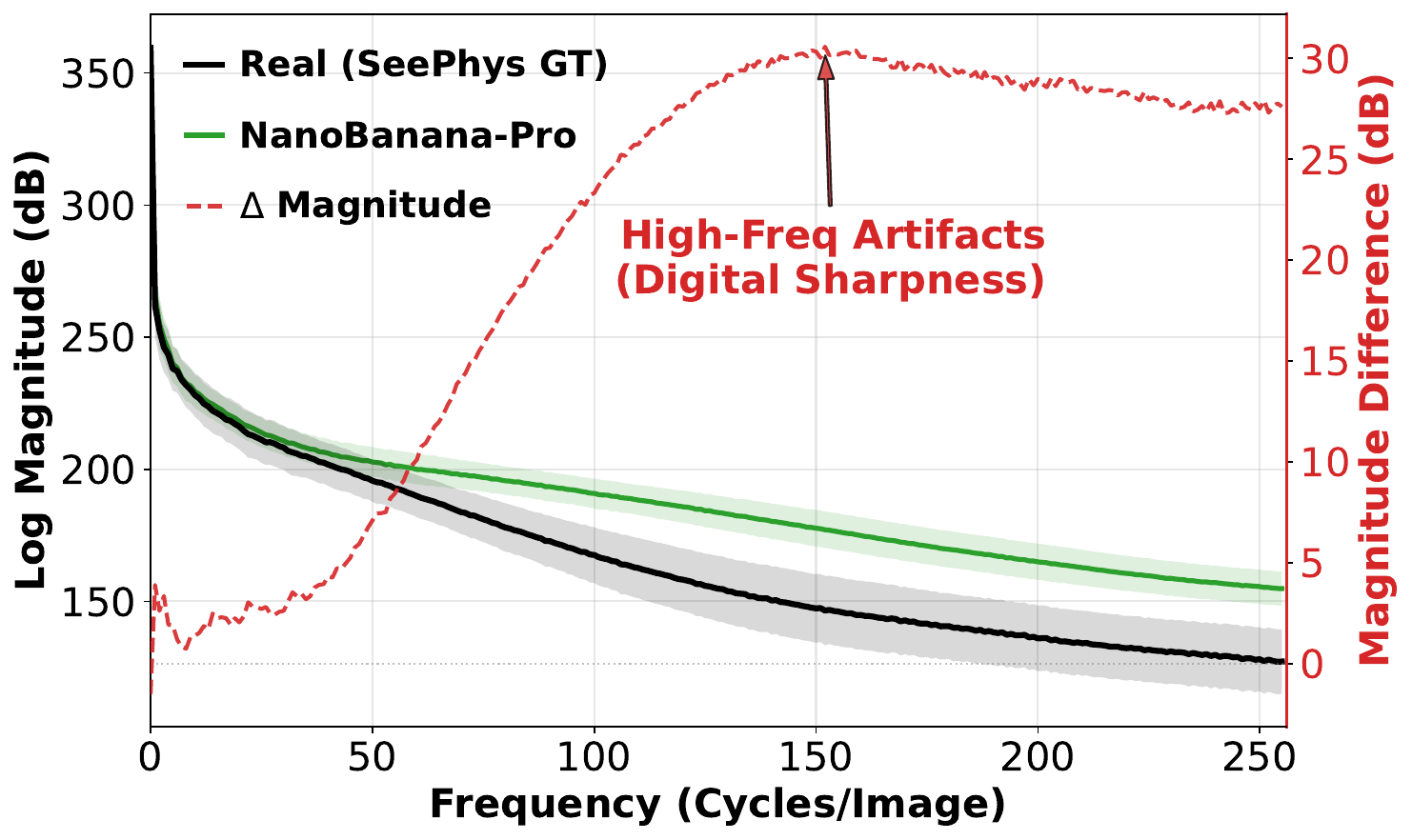}
        \phantomsubcaption
        \label{fig:frequency_nano}
    \end{subfigure}
    \vspace{-20pt}
    
    \caption{\textbf{Analysis of Distributional Gaps.} 
    \emph{Left}: CLIP t-SNE shows clear separation between \texttt{NanoBanana-Pro} and real images. 
    \emph{Right}: Spectral analysis attributes this gap to excessive high-frequency energy in generated images.}
    \label{fig:distributional_gaps}
\vspace{-10pt}
\end{figure*}

\noindent \textbf{Representation Gap.}
We further visualize CLIP embeddings using t-SNE. As shown in Figure~\ref{fig:embedding_nano}, images generated from identical prompts form a distinct, linearly separable cluster (red) from real images (black), indicating a persistent domain gap in visual style despite semantic alignment.

\noindent \textbf{Spectral Bias.}
Spectral analysis (Figure~\ref{fig:frequency_nano}) shows that while low-frequency components align well, \texttt{Nanobanana-Pro} exhibits consistently higher high-frequency energy than real images, with the magnitude difference $\Delta$ peaking in the high-frequency range. This indicates artificial ``digital sharpness'' devoid of natural spectral decay, distinguishing synthetic outputs from real scans. More discussion are provided in Appendix~\ref{apx:gap}.

\subsection{Domain-specific Breakdown}
\label{subsec:domain_breakdown}
\begin{table*}[h]
    \centering
\caption{\textbf{Breakdown of Performance by Subject.}
We report the inverse validation rate ($\mathcal{R}_{\text{inv}}$) and the LMM-as-Judge mean score (Judge) for each discipline.
The last row reports the average performance across all baselines.}

    \label{tab:subject_results}
    \resizebox{\textwidth}{!}{%
    \begin{tabular}{l ccccccccccc}
        \toprule
        \textbf{Model} & 
        \multicolumn{2}{c}{\textbf{Math}} & 
        \multicolumn{2}{c}{\textbf{Physics}} & 
        \multicolumn{2}{c}{\textbf{Chemistry}} & 
        \multicolumn{2}{c}{\textbf{Biology}} & 
        \multicolumn{2}{c}{\textbf{Universal}} \\
        
        \cmidrule(lr){2-3} \cmidrule(lr){4-5} \cmidrule(lr){6-7} \cmidrule(lr){8-9} \cmidrule(lr){10-11}
        
        & 
        \textbf{$\mathcal{R}_{\text{inv}}$} (\%) & \textbf{Judge} & 
        \textbf{$\mathcal{R}_{\text{inv}}$} (\%) & \textbf{Judge} & 
        \textbf{$\mathcal{R}_{\text{inv}}$} (\%) & \textbf{Judge} & 
        \textbf{$\mathcal{R}_{\text{inv}}$} (\%) & \textbf{Judge} & 
        \textbf{$\mathcal{R}_{\text{inv}}$} (\%) & \textbf{Judge} \\
        
        \midrule
        \multicolumn{12}{l}{\textbf{Open-source T2I Models}} \\
        \midrule
        
        HunyuanImage-3.0 &
        13.70 & 0.56 &
        25.39 & 0.94 &
        22.92 & 0.86 &
        33.33 & 1.06 &
        12.58 & 0.68 \\
        
        Qwen-Image &
        30.14 & 0.57 &
        28.12 & 0.85 &
        40.28 & 0.66 &
        43.14 & 0.95 &
        15.23 & 0.62 \\
        
        \midrule
        \multicolumn{12}{l}{\textbf{Closed-source T2I Models}} \\
        \midrule
        
        GPT-Image-1 &
        31.51 & 1.07 &
        32.42 & 1.26 &
        38.19 & 1.20 &
        49.02 & 1.38 &
        29.14 & 1.09 \\

        Flux2-flex &
        41.78 & 0.83 &
        53.52 & 1.18 &
        \underline{60.42} & 1.06 &
        58.82 & 1.17 &
        36.42 & 0.86 \\
        
        Seedream-4.0 &
        42.47 & 0.78 &
        51.56 & 0.98 &
        46.53 & 1.11 &
        41.18 & 1.12 &
        37.09 & 0.82 \\
        
        Nanobanana &
        43.84 & 0.78 &
        51.17 & 1.09 &
        51.39 & 1.00 &
        66.67 & 1.20 &
        31.13 & 0.77 \\
        
        GPT-Image-1.5 &
        52.05 & 1.43 &
        60.16 & 1.61 &
        56.94 & 1.48 &
        72.55 & 1.58 &
        47.68 & 1.41 \\
        
        Nanobanana-Pro &
        64.38 & 1.80 &
        62.50 & 1.84 &
        54.86 & \textbf{1.80} &
        74.51 & \textbf{1.89} &
        60.93 & 1.82 \\

        \midrule
        \multicolumn{11}{l}{\textbf{ImgCoder}} \\
        \midrule
        
        Qwen3-ImgCoder &
        44.52 & 1.54 &
        43.36 & 1.39 &
        52.08 & 1.14 &
        41.18 & 0.96 &
        43.05 & 1.34 \\

        Gemini-3-Flash-ImgCoder &
        \underline{66.44} & \underline{1.89} &
        \underline{70.31} & \underline{1.91} &
        \textbf{63.19} & \underline{1.76} &
        \textbf{80.39} & 1.82 &
        \underline{71.52} & \underline{1.89} \\
        
        Gemini-3-Pro-ImgCoder &
        \textbf{69.86} & \textbf{1.94} &
        \textbf{75.39} & \textbf{1.93} &
        59.03 & \underline{1.76} &
        \underline{76.47} & \underline{1.84} &
        \textbf{72.85} & \textbf{1.91} \\        
        
        \midrule
        \textbf{Average} &
        45.52 & 1.20 &
        50.36 & 1.36 &
        49.62 & 1.26 &
        57.93 & 1.36 &
        41.60 & 1.20 \\
        \bottomrule
    \end{tabular}%
    }
\end{table*}

As shown in Table~\ref{tab:subject_results}, We analyze performance across domains to understand how generation paradigms interact with domain-specific requirements. Mathematics, physics, and universal diagram types demand strict geometric, symbolic, and layout accuracy, where code-based ImgCoder models consistently outperform pixel-based approaches. In contrast, biology and visually rich chemistry subdomains favor pixel-based models, benefiting from organic shapes and dense textures. Chemistry represents a boundary case: molecular structures favor code-based reasoning, while crystal and reaction diagrams are better handled by pixel-based models. Overall, domain performance depends more on structural constraints and information density than on model scale, highlighting the potential of hybrid approaches for scientific image synthesis. Further analyses are provided in Appendix~\ref{apx:domain}.

\begin{figure*}[h]
    \centering
    \begin{subfigure}[c]{0.31\linewidth} 
        \centering
        \includegraphics[width=\linewidth]{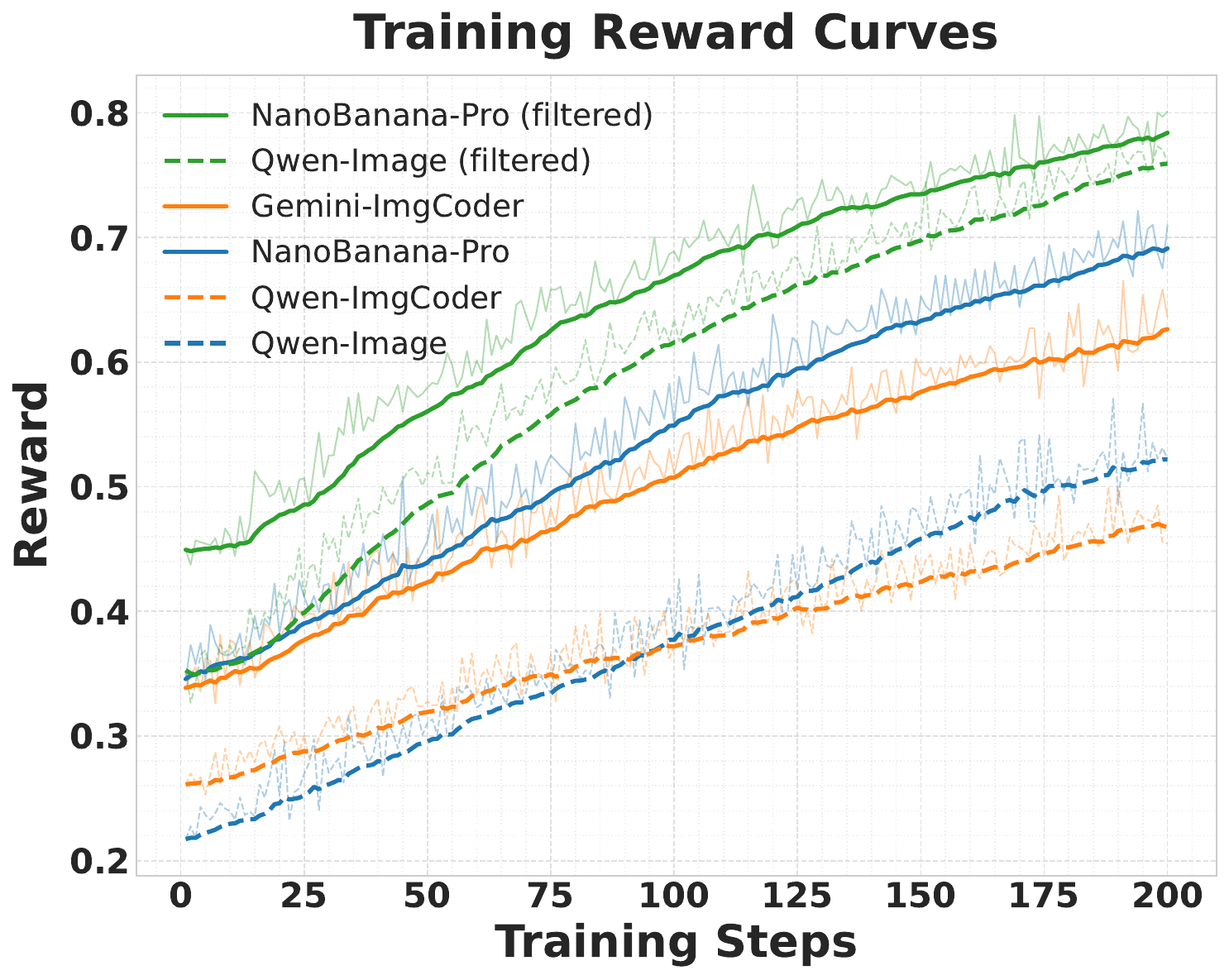}
        \phantomsubcaption 
        \label{fig:reward}
    \end{subfigure}
    \hfill 
    \begin{subfigure}[c]{0.31\linewidth}
        \centering
        \includegraphics[width=\linewidth]{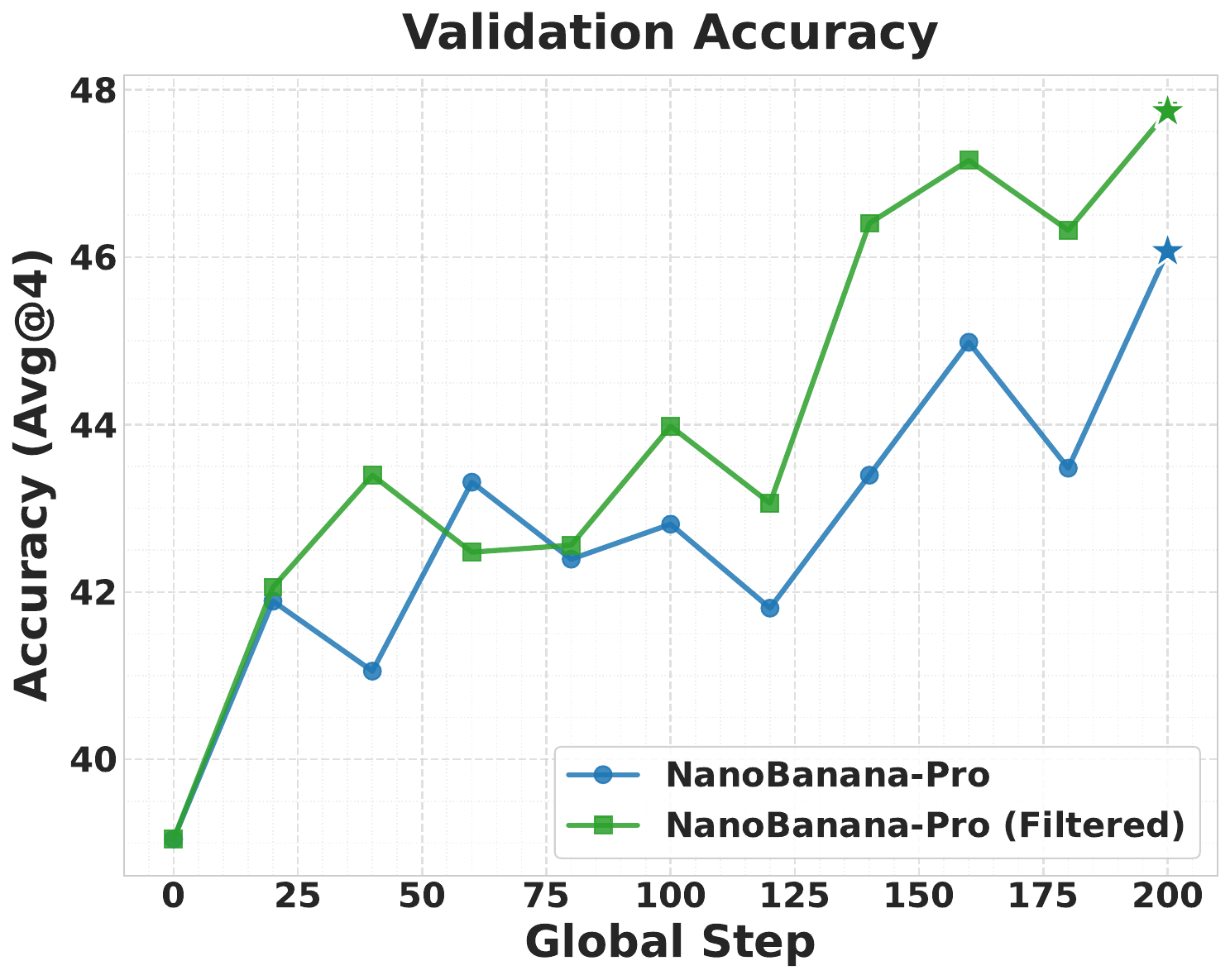}
        \phantomsubcaption
        \label{fig:filtration}
    \end{subfigure}
    \hfill 
    \begin{subfigure}[c]{0.31\linewidth}
    \centering
    \includegraphics[width=\linewidth]{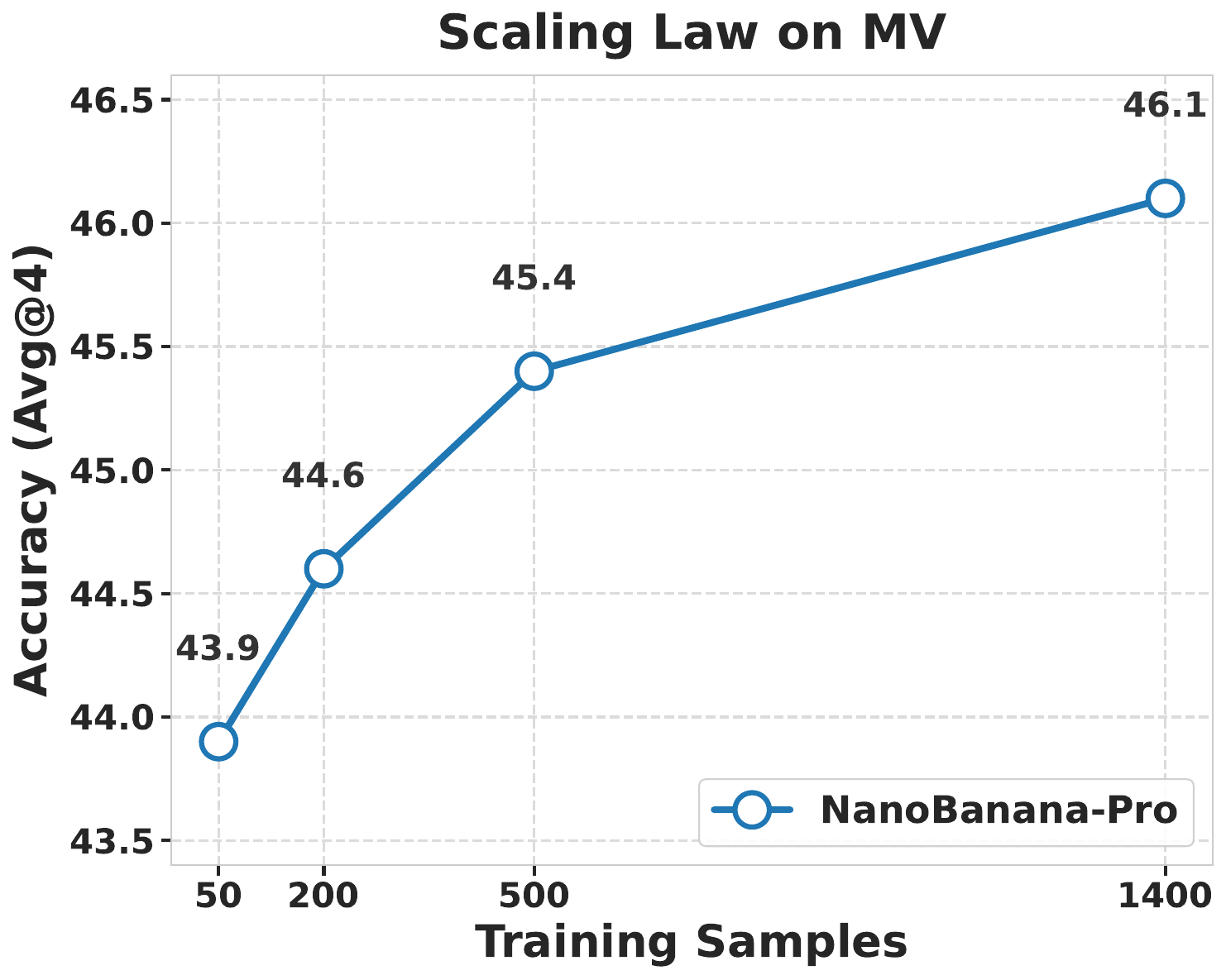}
    \phantomsubcaption 
    \label{fig:scaling}
    \end{subfigure}
    \hfill 
    
    \caption{\textbf{Downstream Data Utility.} \emph{Left (a)}: Stronger teachers yield higher training rewards. \emph{Middle (b)}: Filtered data consistently outperforms unfiltered data. \emph{Right (c)}: Performance scales predictably with data size.}
    \label{fig:rl}
\end{figure*}

\begin{table}[h]
    \centering
\caption{\textbf{Downstream Performance Comparison.}}
        \label{tab:rl_results}
        \resizebox{0.5\linewidth}{!}{%
        \renewcommand{\arraystretch}{0.94}
            \begin{tabular}{lccc}
                \toprule
                \textbf{RL Data} & \textbf{GEO3K} & \textbf{MV} & \textbf{AVG} \\ 
                \midrule      
                Base                    & 61.9 & 39.0          & 54.5 \\
                \midrule
                Qwen-Image           & 68.2          & 45.9          & 57.1 \\
                Qwen-Imgcoder        & 67.9          & 46.9          & 57.4 \\
                Qwen-Image (Filt)    & 68.6          & \underline{47.0}          & 57.8 \\ 
                \midrule
                Nanobanana-Pro      & \textbf{70.1} & 46.1          & \underline{58.1} \\
                Gemini-ImgCoder      & \underline{69.1}          & 46.9          & 58.0 \\
                Nanobanana-Pro (Filt) & 68.7          & \textbf{47.7} & \textbf{58.2} \\
                \bottomrule
            \end{tabular}%
        }%
        \end{table}

\section{Data Utility for Downstream Reasoning}
\label{subsec:downstream}

While Section~\ref{sec:t2i_experiments} assessed intrinsic visual quality, this section evaluate the extrinsic utility of generated images as training data for LMMs. Specifically, we address \textbf{RQ3 - Data Utility}: Whether synthetic scientific images can improve downstream multimodal reasoning via fine-tuning?  Implementation details are in Appendix~\ref{apx:rl_setup}.

\subsection{Effectiveness of Synthetic Image Training} 

As shown in Table \ref{tab:rl_results}, training with synthetic images yields substantial performance improvements over the baseline across all model variants on both GEO3K and MathVision. For instance, the best-performing \texttt{Nanobanana-Pro (Filt)} achieves an average score of 58.2, corresponding to a 3.7-point absolute gain over the baseline (54.5). Consistently, Figure~\ref{fig:reward} shows a steadily increasing reward trend during training, indicating stable and effective optimization dynamics. Notably, ImgCoder variants show a clear advantage on the more challenging and discriminative MV benchmark compared to pixel-based counterparts, indicating that ImgCoder can serve as a low-cost and scalable data synthesis approach with strong structural rigor.

\subsection{Impact of Image Quality}
We further examine the effect of synthetic image quality on downstream performance. Training on higher-quality images leads to better results, with \texttt{Nanobanana-Pro}–based data outperforming \texttt{Qwen-Image} (58.1 vs. 57.1). In addition, applying data filtration consistently improves both training rewards (Figure~\ref{fig:reward}) and test accuracy (Figure~\ref{fig:filtration}), indicating that downstream performance is sensitive to the quality of synthetic images rather than data quantity alone.

\subsection{Scaling Law of Synthetic Data}
To further investigate scalability, we conduct a controlled data-scaling study using synthetic images generated by \texttt{NanoBanana-Pro}. Training sets ranging from 50 to 1.4K samples are constructed. 
As shown in Figure~\ref{fig:scaling}, downstream accuracy exhibits a clear and stable increase with data scale, rising from 43.9\% to 46.1\% (+2.2\%). Notably, the performance curve follows a distinct log-linear growth trend without observable saturation, suggesting that high-fidelity synthetic images can continuously serve as effective training signals. 
These results suggest that high-fidelity synthetic images provide consistently effective supervision signals rather than diminishing returns, and that the well-known “Data Engine” scaling behavior observed in text-domain training extends naturally to multimodal reasoning scenarios.

\section{Conclusion}
In this paper, we investigate scientific image synthesis for multimodal reasoning, revealing a precision–expressiveness trade-off between pixel- and code-based generation. We introduce SciGenBench to systematically evaluate structural correctness and information utility. Experiments show code-based methods like ImgCoder excel in structural precision and reduce failure modes, while pixel-based models offer richer visual expressiveness. Moreover, high-fidelity synthetic images consistently improve downstream multimodal reasoning performance, underscoring the value of scalable, logic-grounded data synthesis for advancing intelligence in LMMs.
\clearpage
\bibliographystyle{mybst}
\bibliography{custom}


\clearpage
\appendix

\section{Experimental Details}

\subsection{Baseline Models Details}
\label{apx:baseline_details}

We provide the detailed citations for the baseline models evaluated in Table~\ref{tab:main_results} below.

\noindent\textbf{Open-source T2I Models.}
We evaluate representative open-source models including \texttt{HunyuanImage-3.0}~\cite{hunyuan} and \texttt{Qwen-Image}~\cite{qwenimage}.

\noindent\textbf{Closed-source T2I Models.}
For proprietary models, we include \texttt{GPT-Image-1} and its upgraded version \texttt{GPT-Image-1.5}~\cite{gptimage}. We also evaluate \texttt{Seedream-4.0}~\cite{seedream}, \texttt{Flux2-flex}~\cite{flux2}, as well as \texttt{Nanobanana} and \texttt{Nanobanana-Pro}~\cite{nanobananapro}.

\noindent\textbf{ImgCoder Backbones.}
Our ImgCoder framework is evaluated using different LLM backbones, specifically \texttt{Qwen3}~\cite{qwen3}, \texttt{Gemini-3-Flash}, and \texttt{Gemini-3-Pro}~\cite{gemini3}.

\subsection{Generative Model Settings}
\label{apx:setup}
All model inferences are conducted using default settings. For closed-source T2I models, we utilize their official APIs. Open-source T2I models are deployed locally on NVIDIA A100 GPUs and implement the official prompt rewriting strategy to ensure optimal performance.

For the ImgCoder framework, we set the sampling temperature to $0.6$. To ensure robustness against syntactic instability, we implement an error-recovery mechanism that allows up to three retries in cases of code extraction failure or compilation errors. If execution remains unsuccessful after these attempts, a blank image is returned as a fallback to maintain pipeline continuity.

\subsection{Downstream Training Configuration}
\label{apx:rl_setup}
We utilize four synthetic datasets in Section~\ref{sec:t2i_experiments}(\texttt{Qwen-Image}, \texttt{Nanobanana-Pro}, \texttt{Qwen-Imgcoder}, \texttt{Gemini-Flash-ImgCoder}) to perform reinforcement learning fine-tuning on the \texttt{Qwen3-VL-8B-Instruct} model. Additionally, we filter out incorrect images from the raw datasets to construct high-quality subsets, specifically \texttt{Nanobanana-Pro (Filt)} and \texttt{Qwen-Image (Filt)}.  

For the training hyperparameters, the model is trained for 200 global steps with a batch size of 128. We perform 8 rollouts per prompt and set the maximum response length to 8192 tokens. All training and evaluation processes are implemented based on the VeRL~\cite{verl} library.  

For evaluation, we select the MathVision\textsubscript{mini}~\cite{mathvision} and Geometry3K\textsubscript{test}~\cite{geo3k} benchmarks. Following the official recommendation for Qwen models, the sampling temperature is set to 0.6. Given the limited sample size of MathVision\textsubscript{mini}, we report the average score over 4 samples (AVG@4) to ensure result stability. To address the challenge of rule-based answer matching in scientific domains, we employ the \texttt{Compass-Verifier-8B}~\cite{compassverifier} as the judge model for both reward computation during training and accuracy assessment during evaluation.

\section{Additional Results}

\begin{figure*}[h]
    \centering
    \begin{subfigure}[b]{0.48\linewidth}
        \centering
        \includegraphics[width=\linewidth]{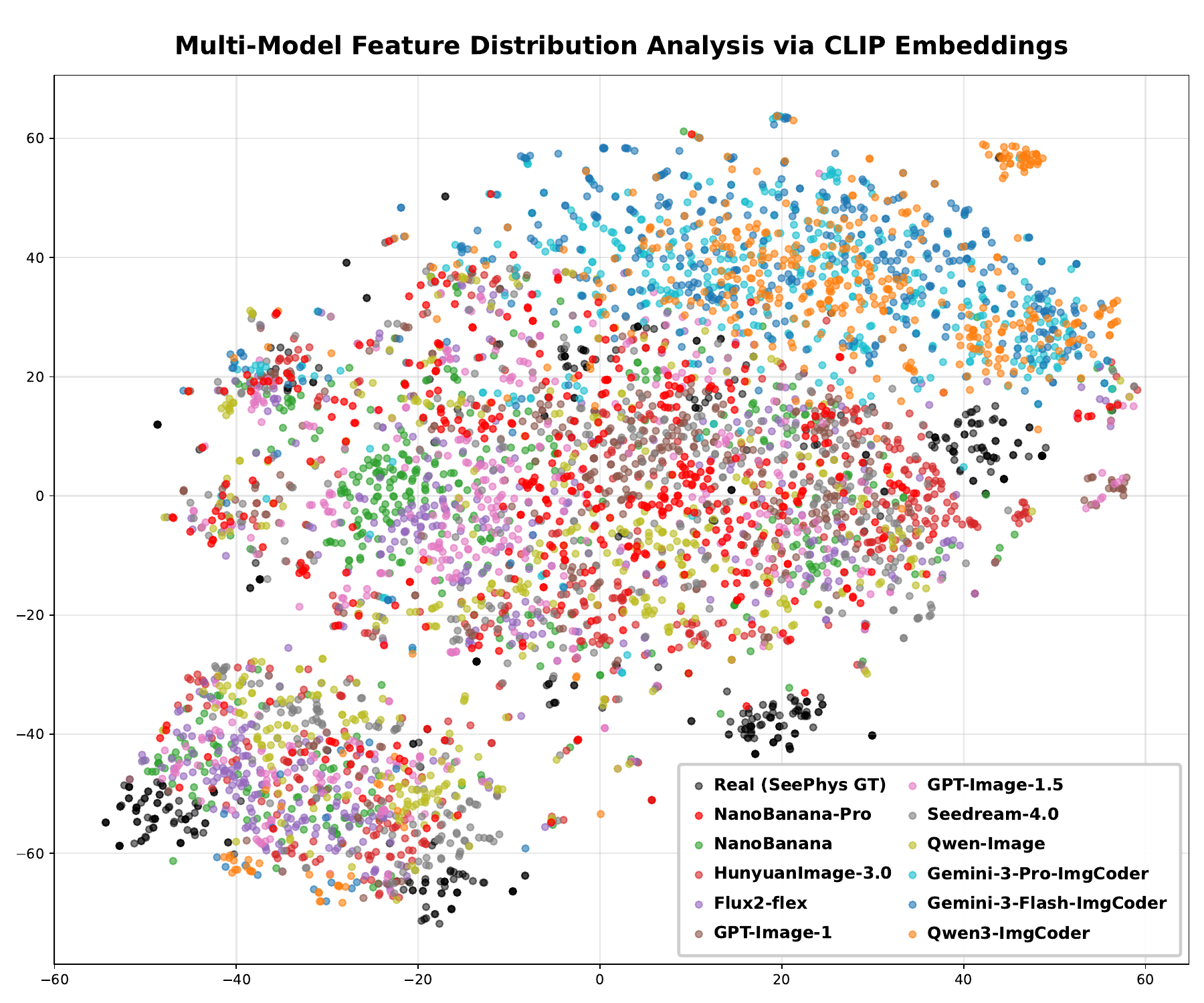}
        \caption{Feature distribution analysis (t-SNE).}
        \label{fig:embedding_all}
    \end{subfigure}
    \hfill 
    \begin{subfigure}[b]{0.48\linewidth}
        \centering
        \includegraphics[width=\linewidth]{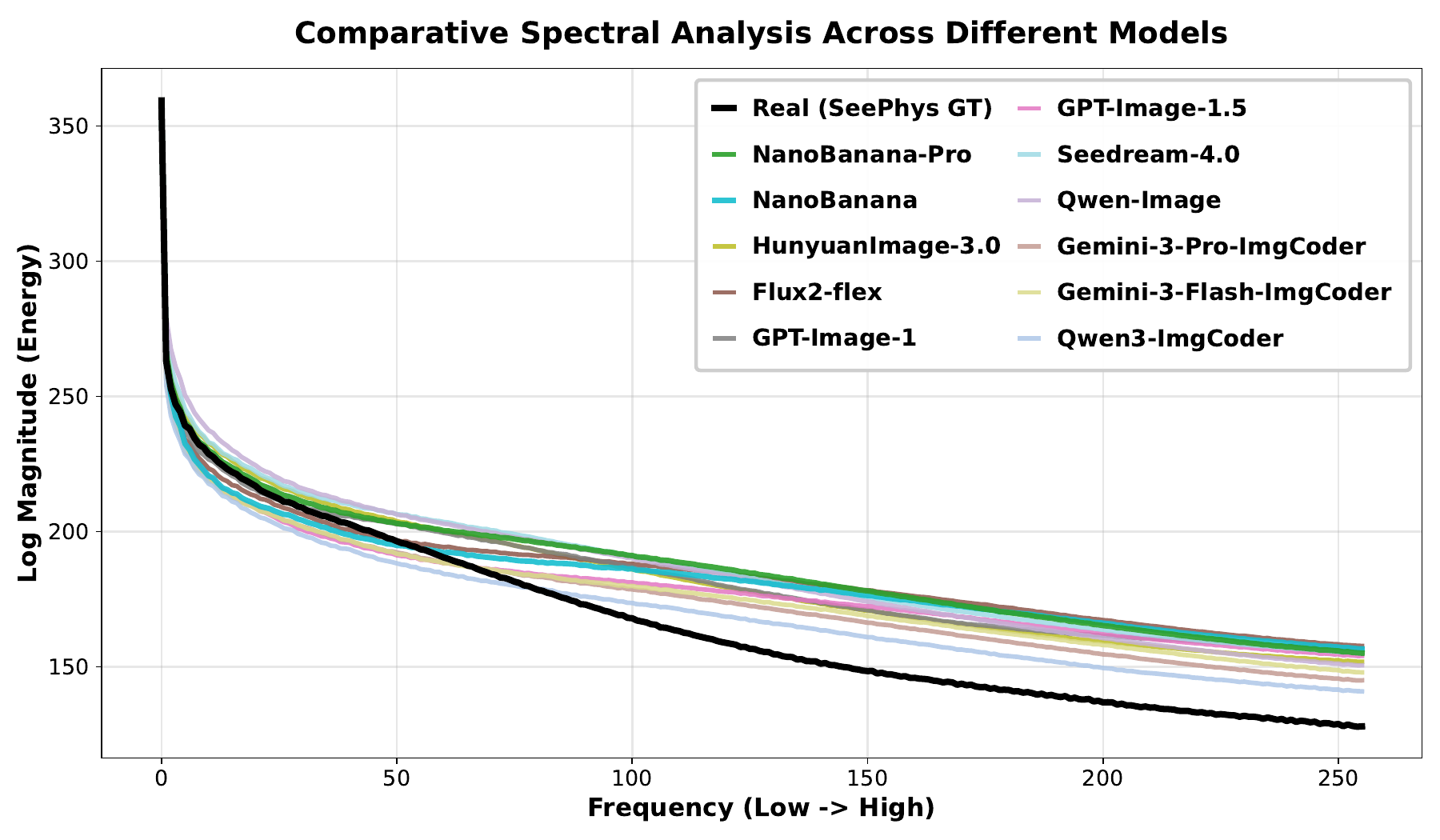}
        \caption{Spectral frequency analysis.}
        \label{fig:freq_all}
    \end{subfigure}
    
    \caption{Systemic domain gap analysis. (a) Visualization of CLIP embeddings shows that real scientific images (black cluster) occupy a distinct region from synthetic ones. (b) Spectral analysis reveals that real images (black curve) have significantly lower high-frequency energy compared to the "digital sharpness" of synthetic models.}
    \label{fig:domain_gap_analysis}
\end{figure*}

\subsection{Distributional Gap with Real Images}
\label{apx:gap}
To validate whether the distributional discrepancy observed in Section~\ref{subsec:dist_gap} is specific to Nanobanana Pro or a broader issue in scientific image generation, we extended our spectral and representation analysis to include all evaluated models.

\noindent \textbf{Universal Representation Gap.} As shown in Figure~\ref{fig:embedding_all}, we visualize the CLIP embedding space for all 11 evaluated models against the real SeePhys dataset. The t-SNE projection reveals a striking pattern: the real scientific images (black points) form a tightly clustered manifold that is distinct from almost all synthetic distributions. The majority of high-fidelity models cluster away from the ground truth. This suggests that current generative paradigms have converged on a "digital scientific style" that is internally consistent but visually distinct from the "natural" distribution of real-world scientific literature.

\noindent \textbf{Consistent Spectral Bias.} Figure~\ref{fig:freq_all} presents the frequency domain analysis across all models. The results confirm that the high-frequency energy divergence is a universal phenomenon. Every single generated model (colored curves) exhibits higher log-magnitude energy in the high-frequency spectrum compared to the real images (black curve). This consistent "spectral gap" indicates that synthetic images lack the natural degradation processes—such as printing imperfections, scanning noise, and paper texture—that characterize real-world scientific data.



\begin{figure*}
    \centering
    \includegraphics[width=1\linewidth]{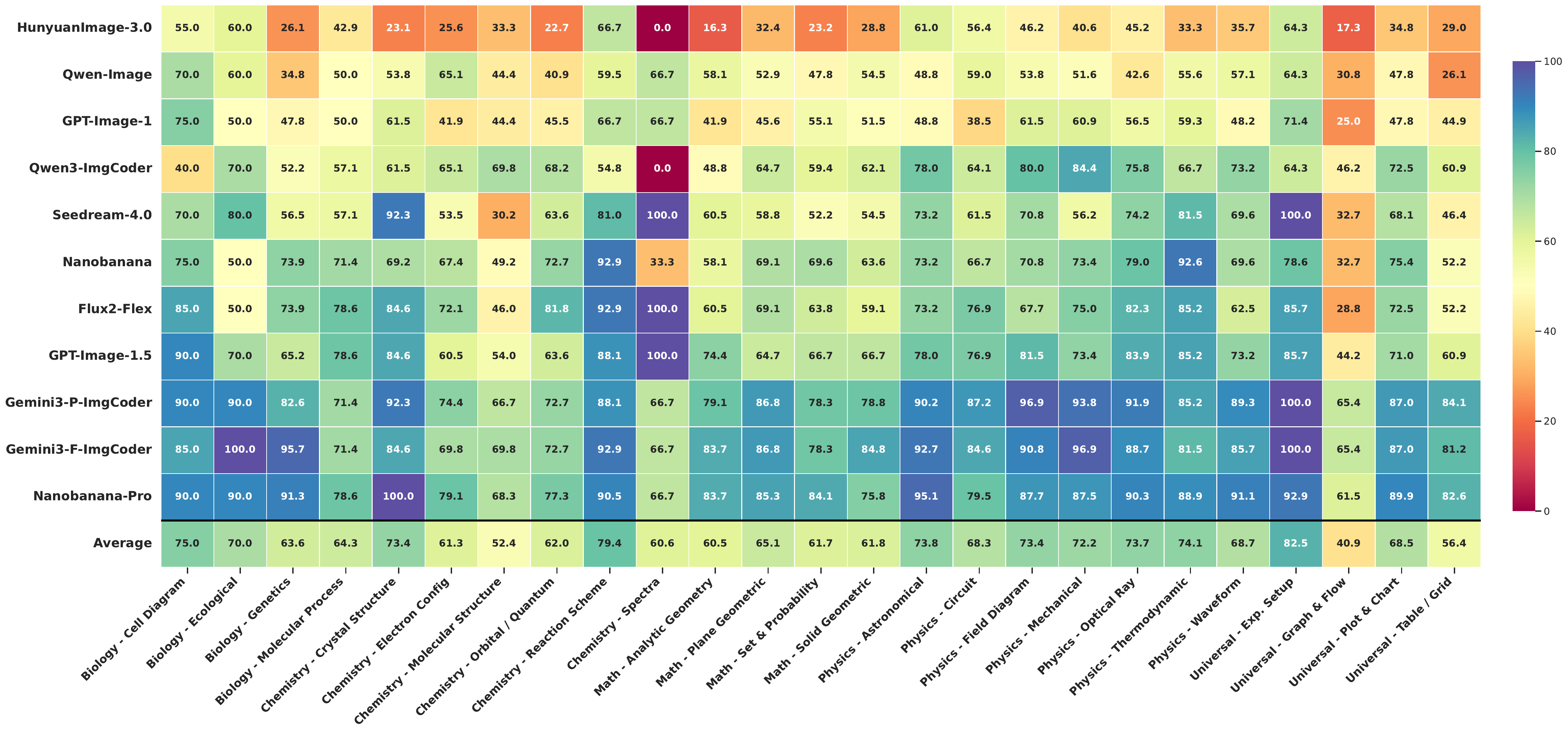}
    \caption{\textbf{Perfect Image Rate by Domain.} The heatmap shows the percentage of strictly correct images (passing all visual validation quizzes) generated by each model across fine-grained scientific sub-categories in the SciGen dataset.}
    \label{fig:heatmap_perf}
\end{figure*}

\begin{figure*}
    \centering
    \includegraphics[width=1\linewidth]{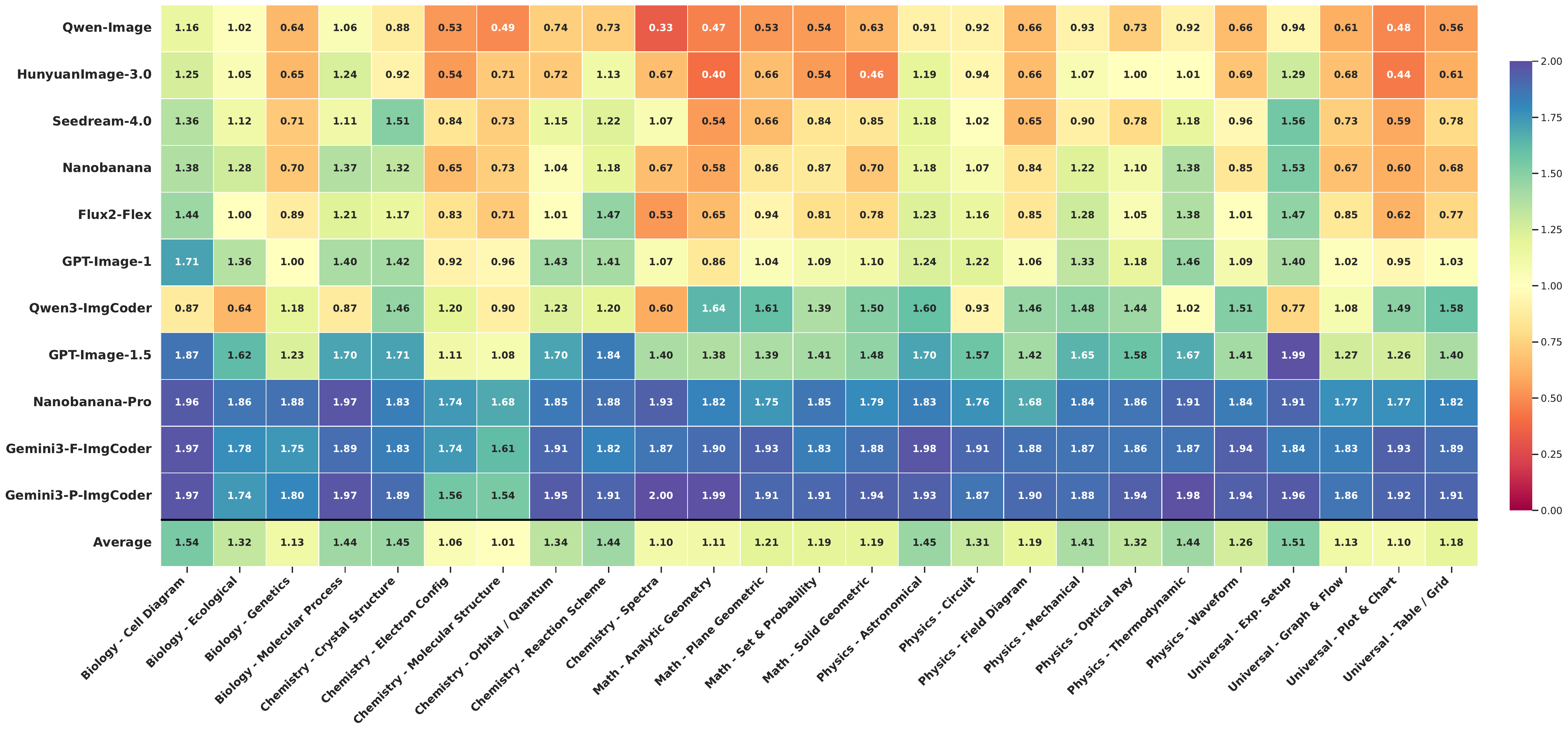}
    \caption{\textbf{Average LMM-Judge Score by Domain.} The heatmap visualizes the mean comprehensive quality score (0--2 scale) awarded by the LMM judge for each model across fine-grained scientific sub-categories.}
    \label{fig:heatmap_judge}
\end{figure*}

\subsection{Domain-specific Breakdown}
\label{apx:domain}
Table~\ref{tab:subject_results}, Figure~\ref{fig:heatmap_perf} and Figure~\ref{fig:heatmap_judge}, reveal clear domain-dependent performance patterns across generation paradigms. Overall, code-based methods consistently dominate in structure-intensive domains, while pixel-based models show relative advantages in visually expressive or loosely constrained scenarios.

\noindent \textbf{Mathematics and Physics.}
In highly structured domains such as mathematics and physics, code-based ImgCoder models significantly outperform pixel-based counterparts. As shown in Table~3, \texttt{Gemini-3-Pro-ImgCoder} achieves the highest inverse validation rates in Math (69.86\%) and Physics (75.39\%), alongside top LMM-as-Judge scores (1.94 and 1.93, respectively). Fine-grained results in Figures~6 and~7 further indicate consistent gains across subcategories involving geometric constructions, field diagrams, optical rays, and analytical plots. These tasks demand strict adherence to geometric constraints, coordinate consistency, and topological correctness—properties naturally enforced by executable code.

\noindent \textbf{Chemistry.}
Chemical domains exhibit a mixed pattern. For abstract and symbol-heavy tasks such as molecular structures and electron configurations, code-based models remain competitive and often superior in Judge scores (Figure~7). However, pixel-based models (e.g., \texttt{Flux2-flex}, \texttt{Nanobanana-Pro}) achieve comparable or slightly higher inverse validation rates in visually rich subdomains such as crystal structures and reaction schematics (Figure~6, Table~3). This suggests that chemistry lies at the boundary between rigid structural reasoning and perceptual pattern recognition.

\noindent \textbf{Biology.}
In biology, pixel-based models demonstrate stronger performance overall. As shown in Table~3, \texttt{Nanobanana-Pro} attains the highest Biology $\mathcal{R}_{\text{inv}}$ (74.51\%) and Judge score (1.89), outperforming most code-based variants. Fine-grained heatmaps (Figure~6) indicate that tasks such as cell diagrams and molecular processes benefit from rich textures, organic shapes, and visual priors learned from natural imagery, where strict geometric constraints are less dominant.

\noindent \textbf{Universal Diagram Types.}
For cross-domain universal diagram categories (e.g., plots, tables, flow charts), ImgCoder regains a clear advantage. Table~3 shows that \texttt{Gemini-3-Pro-ImgCoder} achieves the best Universal performance (72.85\%, Judge 1.91). Figures~6 and~7 further highlight ImgCoder’s robustness in tables, grids, and charts, where precise alignment, spacing, and symbolic accuracy are critical and pixel-based models frequently suffer from dense data collapse.

\noindent \textbf{Summary.}
Taken together, these results indicate that domain performance is primarily governed by the degree of structural constraint and information density. Code-based generation excels in domains requiring explicit computation and strict layout control (Math, Physics, Universal), whereas pixel-based models remain competitive in visually organic domains (Biology) and certain chemistry subfields. This domain-specific complementarity further motivates hybrid and co-evolutionary approaches to scientific image synthesis.

\section{Detailed Taxonomy Definitions}
\label{app:taxonomy}

To construct a structured and comprehensive benchmark, we devised a hierarchical taxonomy comprising 5 primary subject categories and 25 secondary image types. This taxonomy is designed to decouple disciplinary semantics from visual structures. Below, we provide detailed definitions and examples for each category.

\subsection{Universal}
This category encompasses visual structures that are not bound to a specific scientific discipline but are widely used across STEM fields for data representation and logical illustration.

\begin{description}
    \item[Plot \& Chart] 
    This category unifies mathematical function graphs and statistical data charts. It includes visualizations defined on 2D/3D coordinate systems, such as continuous function curves (e.g., $y=f(x)$, parabolas, sinusoids) and discrete data visualizations (e.g., scatter plots, histograms, bar charts). We merged these categories due to their shared reliance on coordinate mapping and similar visual features in scientific contexts.
    
    \item[Graph \& Flow] 
    Visual representations of relationships, hierarchies, or processes consisting of nodes and connecting edges. This includes flowcharts, neural network architectures, decision trees, state transition diagrams, and concept maps.
    
    \item[Table \& Grid] 
    Structured textual or numerical data organized in rows and columns (tables) or regular grid-based layouts (matrices, game grids).
    
    \item[Exp. Setup] 
    Diagrams illustrating generic experimental apparatus or laboratory equipment assemblies that are not strictly limited to a single physics or chemistry domain (e.g., a balance scale, a beaker setup, or a generic measurement device).
\end{description}

\subsection{Mathematics}
\begin{description}
    \item[Plane Geometric] 2D Euclidean geometry problems involving shapes like triangles, circles, polygons, and calculations of area, perimeter, or angles.
    \item[Solid Geometric] 3D geometry problems involving spatial figures such as cubes, spheres, prisms, and cylinders, often focusing on volume or surface area.
    \item[Analytic Geometry] Problems involving geometric shapes defined within a coordinate system, including lines, conic sections (ellipses, hyperbolas), and vector coordinates.
    \item[Set \& Probability] Visualizations related to set theory (Venn diagrams) and probability theory (probability trees, distribution areas).
\end{description}

\subsection{Physics}
\begin{description}
    \item[Mechanical] Diagrams depicting forces, motion, and equilibrium. Common elements include blocks on inclined planes, pulley systems, levers, and free-body diagrams.
    \item[Field Diagram] Visualizations of invisible physical fields, including electric field lines, magnetic field lines, and vector fields representing forces.
    \item[Waveform] Representations of wave propagation, including simple harmonic motion, sound waves, and light wave interference patterns.
    \item[Optical Ray] Geometric optics diagrams illustrating light paths, reflection (mirrors), refraction (lenses), and image formation.
    \item[Astronomical] Diagrams related to celestial mechanics, orbital paths, solar system models, and astrophysical phenomena.
    \item[Circuit] Electrical schematics showing components like resistors, capacitors, inductors, batteries, and switches arranged in series or parallel.
    \item[Thermodynamic] Visualizations of thermodynamic processes, including P-V diagrams, heat engine cycles (e.g., Carnot cycle), and phase change diagrams.
\end{description}

\subsection{Chemistry}
\begin{description}
    \item[Molecular Structure] 2D or 3D representations of chemical molecules, including Lewis structures, ball-and-stick models, and skeletal formulas.
    \item[Reaction Scheme] Diagrams illustrating chemical transformations, including reactants, products, transition states, and reaction mechanisms with electron-pushing arrows.
    \item[Electron Config] Visualizations of electron distribution, such as orbital box diagrams and shell models.
    \item[Crystal Structure] Illustrations of solid-state lattice structures, unit cells, and atomic packing arrangements.
    \item[Spectra] Graphical representations of spectral data, such as NMR, IR, or Mass Spectrometry outputs used for chemical analysis.
    \item[Orbital / Quantum] Visualizations of atomic or molecular orbitals (s, p, d, f orbitals) and quantum mechanical probability clouds.
\end{description}

\subsection{Biology}
\begin{description}
    \item[Cell Diagram] Illustrations of cellular structures, including organelles (mitochondria, nucleus), cell membranes, and microscopic views of tissues.
    \item[Ecological] Diagrams representing ecosystem relationships, such as food webs, food chains, energy pyramids, and biogeochemical cycles.
    \item[Genetics] Visualizations related to heredity, including Punnett squares, pedigree charts, and DNA replication/transcription illustrations.
    \item[Molecular Process] Diagrams depicting biological mechanisms at the molecular level, such as protein synthesis, enzyme catalysis, and signaling pathways.
\end{description}

\clearpage
\onecolumn
\section{Qualitative Examples}
\label{apx:cases}

\begin{figure*}[h]
    \centering
    \includegraphics[width=1\linewidth]{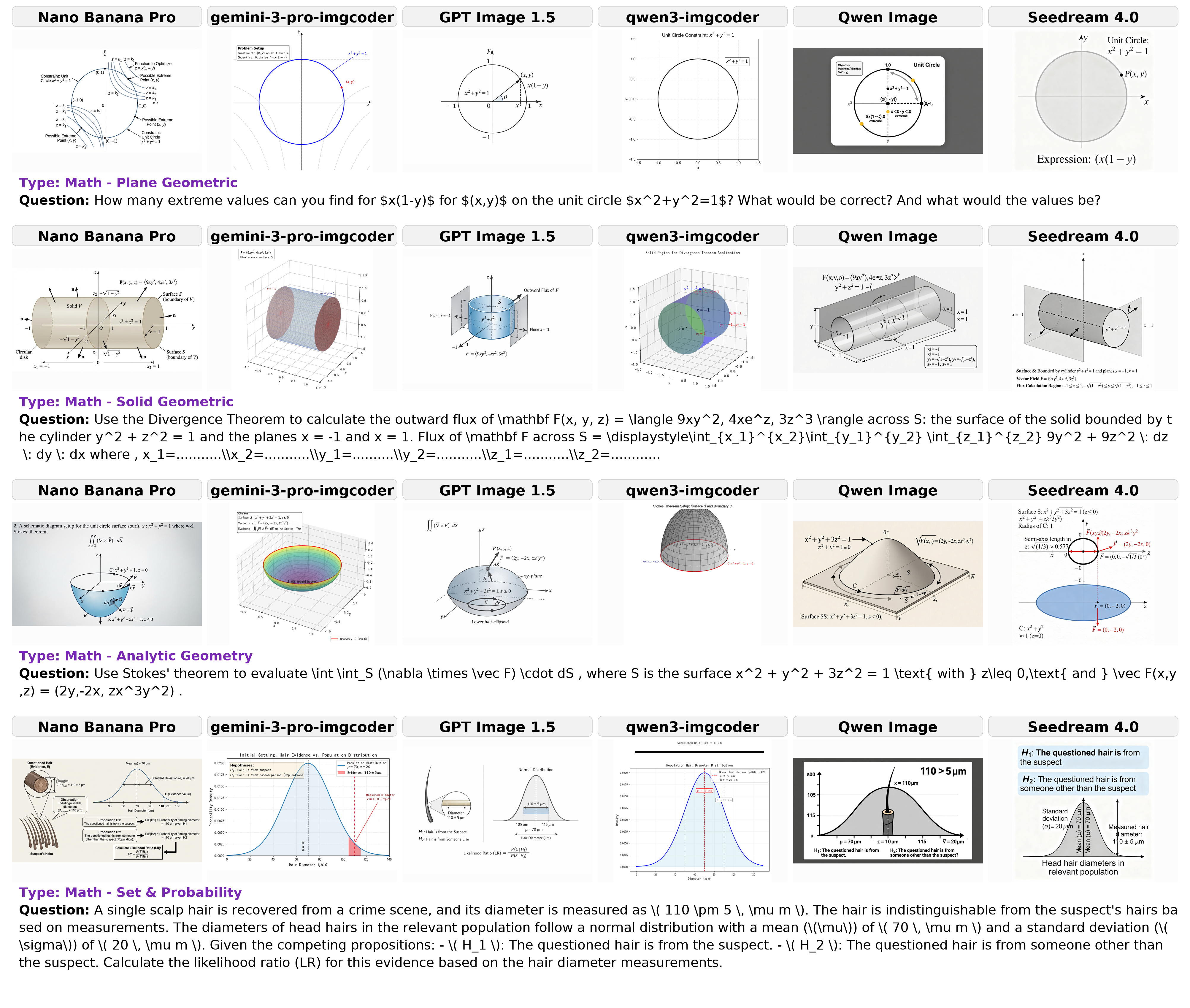}
    \caption{Qualitative comparison in Math problems.}
    \label{fig:qual_math}
\end{figure*}

\begin{figure*}
    \centering
    \includegraphics[width=1\linewidth]{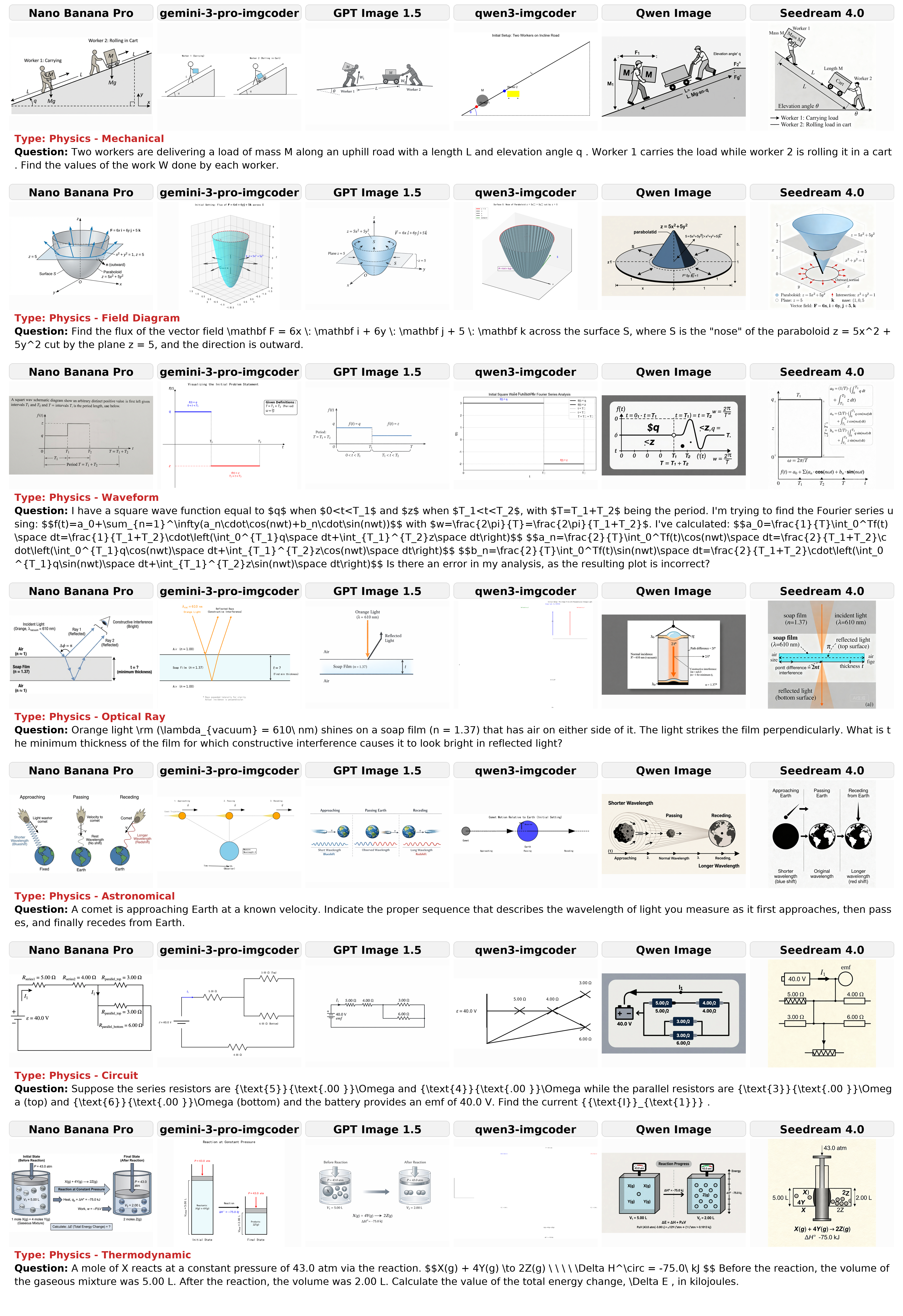}
    \caption{Qualitative comparison in Physics problems.}
    \label{fig:qual_physics}
\end{figure*}

\begin{figure*}
    \centering
    \includegraphics[width=1\linewidth]{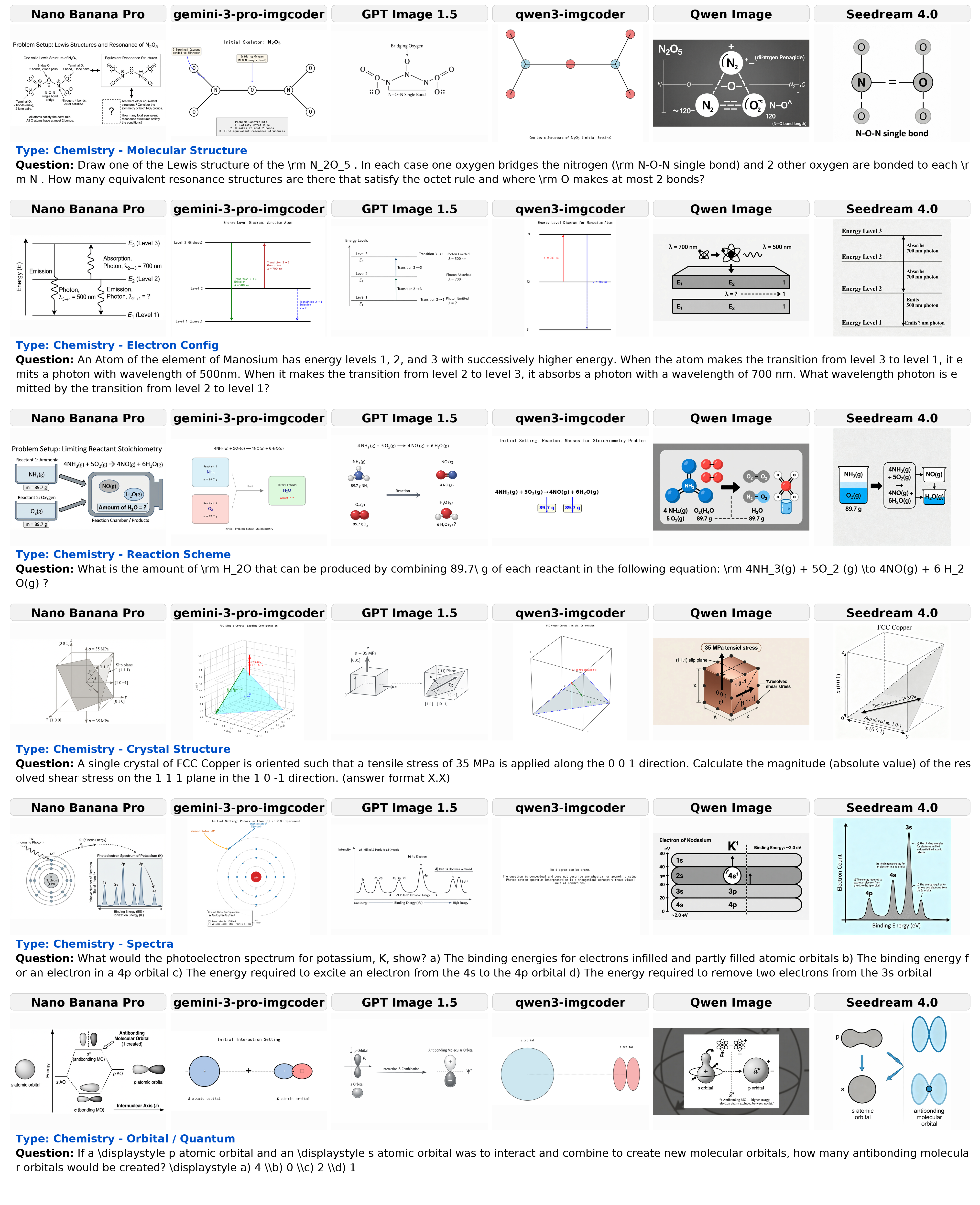}
    \caption{Qualitative comparison in Chemistry problems.}
    \label{fig:qual_chemistry}
\end{figure*}

\begin{figure*}
    \centering
    \includegraphics[width=1\linewidth]{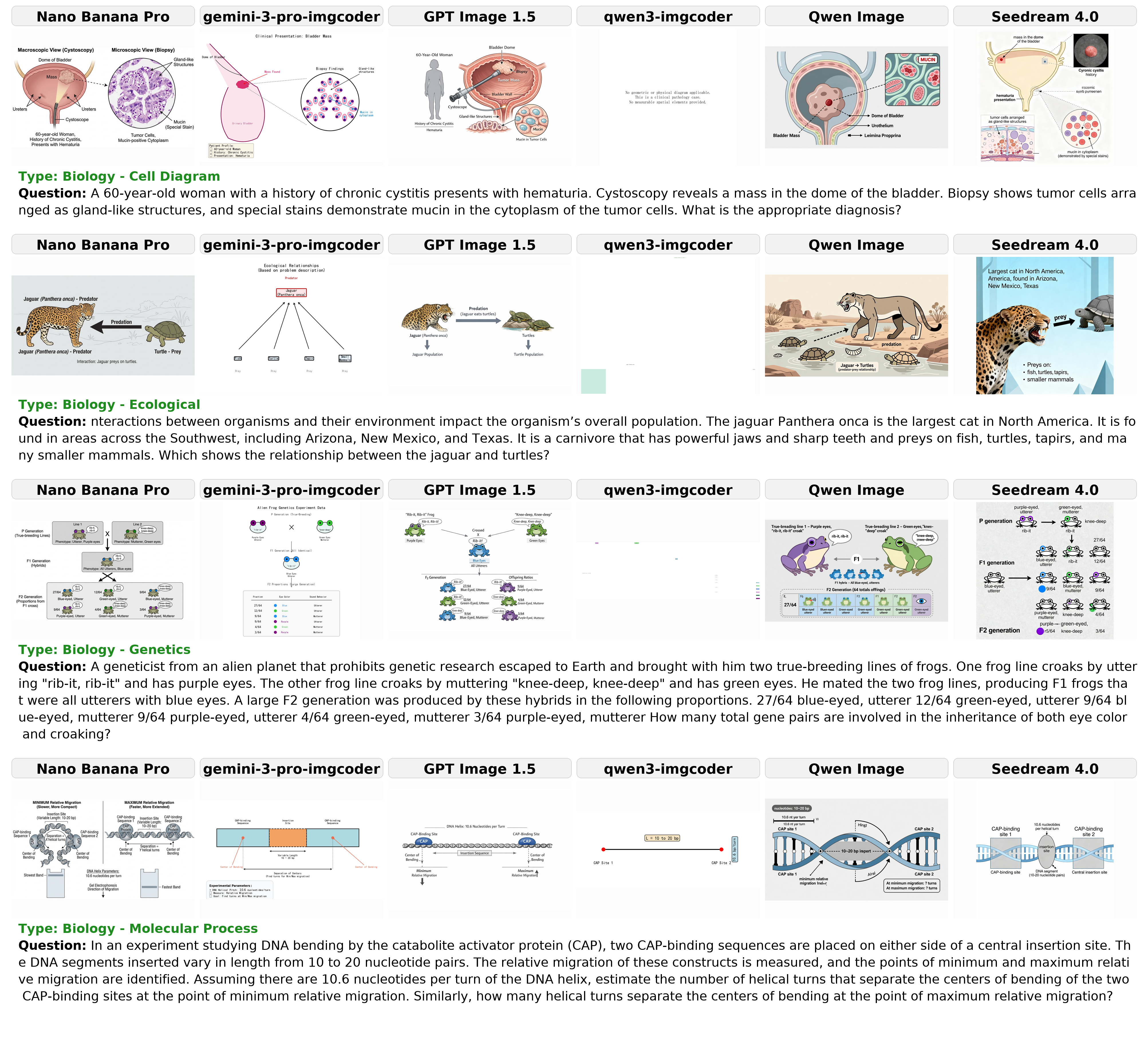}
    \caption{Qualitative comparison in Biology problems.}
    \label{fig:qual_biology}
\end{figure*}

\begin{figure*}
    \centering
    \includegraphics[width=1\linewidth]{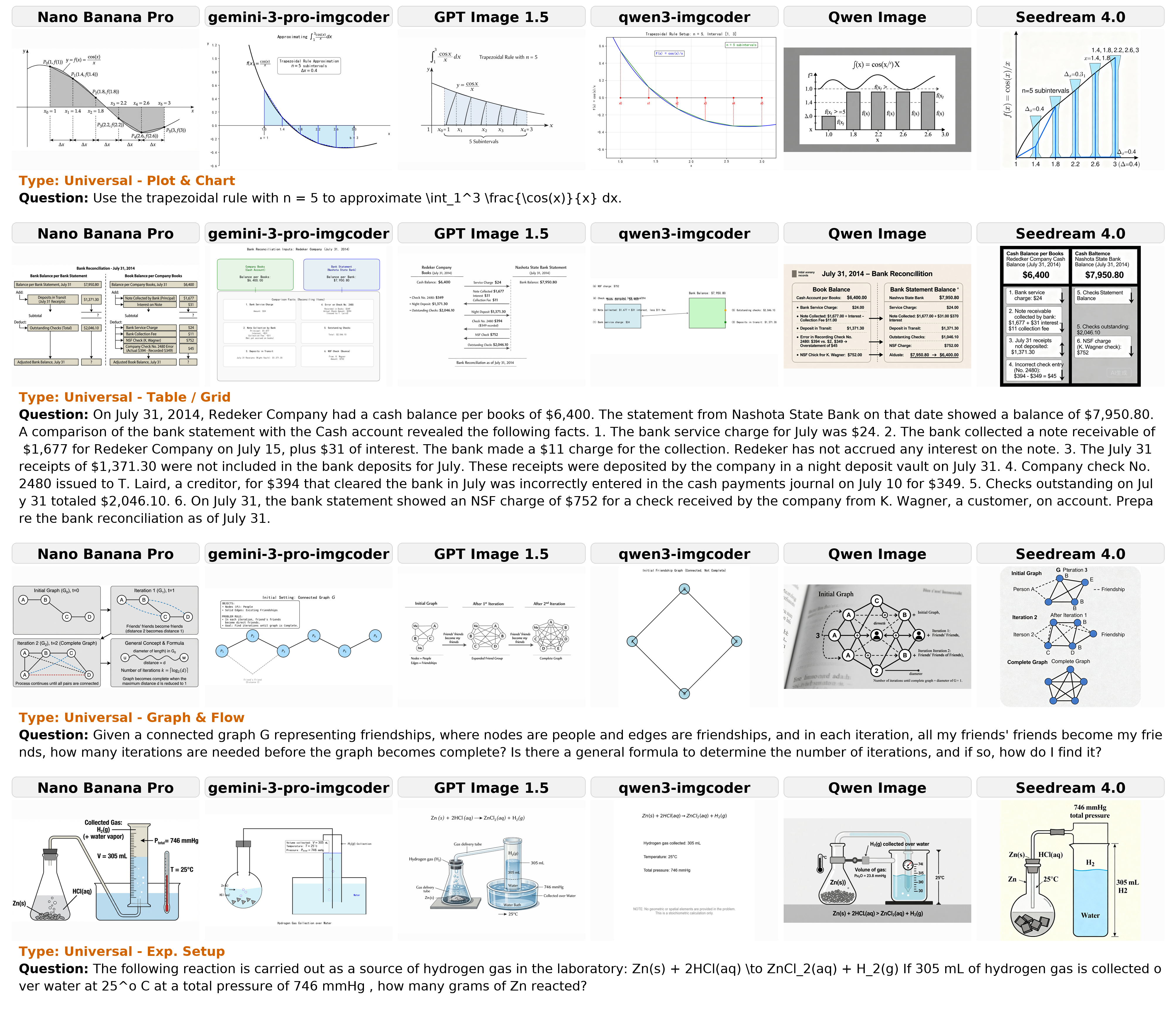}
    \caption{Qualitative comparison in Universal problems.}
    \label{fig:qual_universal}
\end{figure*}

\clearpage
\onecolumn
\section{Prompts}
\label{apx:prompts}
\begin{figure*}[h] 
\begin{promptbox}{Prompt: STEM Text-to-image Generation}
You are a professional STEM problem illustrator.

Your task is to generate a clear, precise, and information-complete diagram that visually represents the scientific scenario described in the problem below.

The illustration must:

- Faithfully reflect the problem context and scenario.

- Explicitly depict all entities, objects, variables, and conditions mentioned in the problem.

- Encode quantitative or relational information visually when possible (e.g., distances, angles, forces, directions, labels, axes).

- Help a student understand the setup of the problem, not the solution.

The illustration must NOT:

- Introduce any assumptions, values, or objects not stated or directly implied by the problem.

- Include solution steps, calculations, or conclusions.

- Add decorative or artistic elements unrelated to the problem.

Use a clean, textbook-style schematic:

- Neutral colors

- Clear labels and annotations

- Simple geometric shapes or standard scientific symbols

**Problem Text:**

\{question\}
\end{promptbox}
\caption{The system prompt used for STEM text-to-image generation.}
\label{pmt:t2igen}
\end{figure*}

\twocolumn

\begin{figure*}[h]
\footnotesize
\begin{promptbox}{Prompt: ImgCoder}
You are a multimodal reasoning assistant skilled in scientific visualization using Python.

You will be given:

- **Original Question** (`{{question}}`): The complete problem text.

Your task:

First, carefully **PLAN** the diagram (reasoning stage).
Then, produce a standalone, runnable Python script using Matplotlib (coding stage).

**Key Objective**: Create a diagram that fully represents the **Initial Setting** of the problem.
   
   - **Completeness**: Visualize all physical objects, geometric shapes, and **given values** (e.g., lengths, angles, forces, labels) mentioned in the text.
   
   - **Confidentiality**: Do NOT reveal the final answer, result, or derivation steps. Only show what is *given* before the problem is solved.
Always use a clear **textbook-style illustration style**.

\#\#\# Output Format (two sections, clearly separated)

\#\#\#\# **Section 1: Plan**

Provide a structured reasoning plan containing the following **four parts**:

1. **Image Content** — Describe the elements to be drawn (shapes, objects, points, lines). Ensure all entities mentioned in the text are accounted for.

2. **Layout** — Explain the approximate spatial arrangement: relative positioning, coordinates estimation, scale.

3. **Labels** — Specify labels and annotations. **Crucial**: List all **given values** (numbers, variables) from the text that must be labeled on the diagram to show the initial state.

4. **Drawing Considerations** — Mention stylistic or logical constraints:

   - What must **NOT** be shown (to avoid solution leakage).
   
   - Matplotlib\-specific details (e.g., `set\_aspect('equal')`, patches).

Each section should be 1–4 bullet points.

---

\#\#\#\# **Section 2: Python Code**

Provide a **complete and runnable Python script**, formatted like this:

\begin{lstlisting}[language=Python, basicstyle=\linespread{0.7}\ttfamily\scriptsize]
```python
import matplotlib.pyplot as plt
import matplotlib.patches as patches
import numpy as np

def draw_diagram():
    # 1. Setup Figure
    fig, ax = plt.subplots(figsize=(6, 6))
    ax.set_aspect('equal') # Crucial for geometry/physics
    ax.axis('off')         # Hide axes unless strictly necessary for graphs

    # 2. Define Coordinates
    # ...

    # 3. Draw Elements (Shapes, Lines, etc.)
    # ...

    # 4. Add Labels and Annotations
    # (Ensure all given values from the question are visibly labeled)
    # ...

    # 5. Finalize and Show
    plt.tight_layout()
    plt.show()

if __name__ == "__main__":
    draw_diagram()
````
\end{lstlisting}

**Rules:**

  * **Initial Setting Only**: The diagram must represent the problem *before* any solution steps are taken. Visualize the "Given" but hide the "Solution".
  
  * **Completeness**: If the text says "radius is 5", the diagram must show the circle and label the radius "r=5".
  
  * **Geometric Accuracy**: Ensure `ax.set\_aspect('equal')` is used to prevent distortion for geometric/physical diagrams.
  
  * **Style**: Textbook aesthetic—consistent line thickness, clean alignment, clear font sizes.

-----

\#\#\# **Input Fields**

Original Question:

\{question\}

Now, follow the format strictly and generate the output.
\end{promptbox}
\caption{The system prompt used for Imgcoder method.}
\label{pmt:imgcoder}
\end{figure*}

\begin{figure*}[h]
\footnotesize
\begin{promptbox}{Prompt: STEM Data Curation (Filtering \& Taxonomy)}
You are an expert STEM data curator and multimodal content analyzer. Your task is to process a raw text-based STEM problem to determine its suitability for "Text-to-Image" synthesis. 

If the problem is valid, you must classify it into a visual taxonomy and assess its quality based on scene clarity and visual complexity.

Input Data:

- Subject: \{subject\}

- Question: \{question\}

---

\#\#\# Step 1: Data Filtering \& Visualizability Check

Analyze the question text. Set `is\_valid` to `false` if ANY of the following "Dirty Data" criteria are met:

1. **Dependency on Missing Context:**
   - The question references a missing image (e.g., "As shown in the figure...", "Which label in the diagram...").
   
2. **Missing Essential Data:**
   - The question asks for a calculation but lacks specific numbers/equations (e.g., "Calculate the probability given the provided values..." but values are missing).
   
   - General theoretical discussion without a specific scenario.
   
3. **Wrong Task Type:**
   - Pure proof, code writing, or instructions to "draw a graph" (we need descriptions, not commands).
   
4. **Non-Visual / Abstract:**

   - Concepts too abstract to be represented by a scientific diagram.

---

\#\#\# Step 2: Taxonomy Classification (Only if `is\_valid` is true)

Classify the image type based on the `{subject}` and visual nature.

*Use 'Universal' if the visual structure is generic (e.g., charts, graphs) and not unique to the subject.*

**Taxonomy:**
1. **Math:** Plane Geometric, Solid Geometric, Analytic Geometry, Set \& Probability

2. **Physics:** Mechanical, Field Diagram, Waveform, Optical Ray, Astronomical, Circuit, Thermodynamic

3. **Chemistry:** Molecular Structure, Electron Config, Reaction Scheme, Crystal Structure, Spectra, Orbital / Quantum

4. **Biology:** Cell Diagram, Ecological, Genetics, Molecular Process

5. **Universal:** Function Graph, Table / Grid, Data Chart, Node-Link, Exp. Setup

---

\#\#\# Step 3: Quality Assessment (Only if `is\_valid` is true)

Rate the following two metrics on a scale of 1 to 5.

**1. Scene Clarity Score (1-5): How explicit is the visual description?**

* **1 (Vague):** Ambiguous scenario. Hard to draw without hallucinating details (e.g., "A car moves on a road" - no speed, angle, or surroundings specified).

* **3 (Moderate):** Key elements are present, but some layout details need inference (e.g., "A triangle with sides 3 and 4" - angle or orientation implied but not stated).

* **5 (Crystal Clear):** Fully deterministic. All coordinates, values, labels, and spatial relationships are explicitly stated (e.g., "A circle centered at (0,0) with radius 5, intersecting a line y=2").

**2. Visual Complexity Score (1-5): How dense/complex is the resulting image?**
* **1 (Simple):** Single object or very simple relationship (e.g., one standalone chemical bond, a single rectangle).

* **3 (Medium):** Multiple interacting components (e.g., a pulley system with 2 blocks, a benzene ring).

* **5 (Complex):** Highly dense information, intricate topology, or many distinct entities (e.g., a complex food web, a circuit with mixed parallel/series resistors and bridges, a detailed eukaryotic cell structure).

---

\#\#\# Output Format
Provide your response in JSON format only.
\begin{lstlisting}[basicstyle=\linespread{0.75}\ttfamily\scriptsize]
```json
{
  "reasoning": "Briefly explain validity decision, taxonomy choice, and score justification.",
  "is_valid": true/false,
  "primary_category": "Math/Physics/Chemistry/Biology/Universal or null",
  "secondary_type": "Specific Type or null",
  "scene_clarity_score": 1-5 (Integer) or null,
  "visual_complexity_score": 1-5 (Integer) or null
}
\end{lstlisting}
\end{promptbox} 
\caption{The system prompt used for visualizability filtering and taxonomy classification.} 
\label{pmt:curation} 
\end{figure*}

\begin{figure*}[h]
\footnotesize
\begin{promptbox}{Prompt: Visual Quiz Generator}
You are a precision-focused Dataset Generator for evaluating Large Multimodal Models. Your goal is to convert scientific text descriptions into a structured JSON object containing a visual checklist and a validation quiz.

**Task:**

Analyze the provided `[Input Text]` to:

1.  Extract a comprehensive **Visual Checklist** of atomic facts that *must* be visualized.

2.  Generate a **Validation Quiz** to verify these visual details without relying on external knowledge.

**Input Text:**

\{question\}

**Phase 1: Visual Checklist Generation (`elements`)**

* **Goal:** Create a list of "Atomic Visual Constraints" that a validator would check off one by one.

* **Requirements:**

    * Each element must be a specific, standalone visual fact derived from the text.
    
    * Include **Values** (e.g., "Label '50kg'"), **Relationships** (e.g., "Box A is on top of Box B"), **Attributes** (e.g., "Dashed line for the normal force"), and **Directions** (e.g., "Arrow points to the right").
    
    * **Format:** Short, descriptive strings (e.g., "Resistor R1 = 10k$\omega$", "Current I flows clockwise").

**Phase 2: Quiz Generation (`quiz`)**

* **Goal:** Create Multiple-Choice Questions (MCQs) that force a model to *look* at the image to verify the checklist items.
* **Anti-Cheating Rules:**

    1.  **Instance-Specific ONLY:** Do not ask general knowledge questions (e.g., "What is gravity?"). Ask only about the specific values/labels described (e.g., "What value is labeled for gravity in this diagram?").
    
    2.  **Visual Verifiability:** Phrasing should imply visual inspection (e.g., "What is the direction of the arrow labeled 'F'?").
    
    3.  **Hard Negatives:** Use *other* numbers/entities from the text as distractors to test precise visual grounding.
    
\vspace{0.1cm}
**Output Format (Strict JSON):**

Return a single JSON object. Ensure the JSON is valid and parsable.

**JSON Schema:**
\begin{lstlisting}[basicstyle=\linespread{0.75}\ttfamily\scriptsize]
```json
{{
    "elements": [
        "String describing visual fact 1 (e.g., 'Resistor R1 is labeled 100 ohm')",
        "String describing visual fact 2 (e.g., 'Vector F points at 45 degrees')",
        "String describing visual fact 3 (e.g., 'Point A is connected to Point B')"
    ],
    "quiz": [
        {{
            "question": "The question string (visually grounded)",
            "options": {{
                "A": "Option text",
                "B": "Option text",
                "C": "Option text",
                "D": "Option text"
            }},
            "correct_option": "A",
            "evidence_snippet": "Substring from text proving this fact"
        }}
    ]
}}
\end{lstlisting}

\end{promptbox} 
\caption{The system prompt used for constructing visual validation quizzes.} 
\label{pmt:quizgen} 
\end{figure*}

\begin{figure*}[h]
\scriptsize
\begin{promptbox}{Prompt: LMM-as-Judge Evaluator}
You are an expert evaluator of scientific and technical diagrams (e.g., geometry, physics, chemistry).

Evaluate the image against the caption on these 5 dimensions:

\#\#\# 1. Correctness \& Fidelity (0--2)

Core Question: Does the image completely and accurately represent all elements, labels, and spatial/logical relationships from the caption, with no omissions OR hallucinations?

* **2 (High):** Perfect match. All elements (points, lines, shapes, labels) from the caption are present and correct. All specified spatial (e.g., 'left of', 'inside') and logical (e.g., 'perpendicular', 'tangent', 'connected to') relationships are perfectly accurate. Crucially, there are NO spurious or "hallucinated" elements (e.g., random lines, meaningless intersections) not implied by the caption.

* **1 (Medium):** Mostly correct. Most key elements are present, but with minor omissions, misplacements, or simplifications. Spatial/logical relationships are mostly right but have slight inaccuracies. May have minor spurious elements that don't confuse the main subject.

* **0 (Low):** Major mismatch. Key elements are missing, incorrect, or relationships are wrong. Or, the image contains significant spurious content (visual noise, random intersections) that contradicts or confuses the caption.

---

\#\#\# 2. Layout \& Precision (0--2)

Core Question: Is the layout clear and technically precise? Does the visual arrangement correctly reflect the logical coordinates and relative spatial positions described?

* **2 (High):** Professional and spatially accurate. Layout is clear, balanced, and precise (straight lines, exact connections). Visual positions perfectly match the logical labels/coordinates (e.g., $(10,0)$ is distinctively right of $(2,0)$) and relative positions are strictly maintained.

* **1 (Medium):** Generally readable but with minor distortions. Layout is understandable but may have slight alignment issues or imprecision. Relative positions are mostly correct (topology preserved), but scale or visual distances may be inaccurate (e.g., order is right, but proportions are off).

* **0 (Low):** Sloppy or spatially contradictory. Layout is cluttered, chaotic, or elements are poorly proportioned. Lines are visibly imprecise/disconnected, OR elements are placed in positions that contradict their coordinates (e.g., positive coordinates drawn on the negative axis, or inverted positions).

---

\#\#\# 3. Readability \& Occlusion (0--2)

Core Question: Do visual elements or labels overlap or occlude each other in a way that obscures meaning or reduces readability?

* **2 (High):** No occlusion. Every element (shapes, arrows, text labels) is fully distinct and clearly separated, with no confusing overlap.

* **1 (Medium):** Minor overlap. Some elements or labels slightly touch or overlap, but it only marginally affects readability (e.g., an arrowhead just touches a label). The core content remains understandable.

* **0 (Low):** Significant occlusion. Key elements or labels overlap heavily, making parts of the diagram unreadable, ambiguous, or indistinguishable.

---

\#\#\# 4. Scientific Plausibility (0--2)

Core Question: Does the image visually conform to the basic principles and conventions of its scientific domain (e.g., physics, geometry), even if not explicitly stated in the caption?

* **2 (High):** Visually plausible. The image "looks right" for its domain. E.g., geometric angles/proportions look reasonable; physics vectors (if representing equilibrium) look balanced; chemical bond angles appear conventional (e.g., VSEPR).

* **1 (Medium):** Minor implausibility. The image is scientifically/logically functional but has minor visual flaws (e.g., a 90° angle looks like 80°; a molecule's bond angle is visibly awkward but still conveys the connection).

* **0 (Low):** Visually implausible. The image clearly violates basic scientific/logical principles in its visual representation (e.g., a force diagram that is obviously unbalanced; a geometric proof figure that is impossibly skewed).

---

\#\#\# 5. Expressiveness \& Richness (0--2)

Core Question: Does the image completely and vividly reproduce the scenario described in the problem?

* **2 (High):** Comprehensive reproduction. The image not only contains the correct elements but also effectively conveys the full *context* or *situation* of the problem. It is visually rich and fully illustrates the prompt's intent.

* **1 (Medium):** Basic representation. The image depicts the necessary elements for the problem but lacks contextual richness or detail. It is functional but minimal.

* **0 (Low):** Incomplete scenario. The image fails to convey the setting or context of the problem, making it difficult to understand the "story" or situation behind the diagram.

---

\#\#\# **Output Format**

Provide short reasoning for each dimension, then output a JSON object with integer scores.

**Example Output:**

**Reasoning:**

* **Correctness \& Fidelity:** The image correctly shows all 5 points and the 3 lines connecting them as described. All labels are present. No extra lines appear.

* **Layout \& Precision:** Lines are straight and connect perfectly at the nodes. The layout is balanced.

* **Readability \& Occlusion:** Label 'A' and 'B' are slightly too close, but do not overlap. All elements are readable.

* **Scientific Plausibility:** The diagram (a geometric proof) shows angles that appear consistent with the "given" perpendicular lines.

* **Expressiveness \& Richness:** The diagram fully captures the geometry problem's scenario, clearly visualizing the intersecting planes described in the text.

\begin{lstlisting}[basicstyle=\linespread{0.75}\ttfamily\scriptsize]
JSON
{
  "Correctness_Fidelity": 2,
  "Layout_Precision": 2,
  "Readability_Occlusion": 2,
  "Scientific_Plausibility": 2,
  "Expressiveness_Richness": 2
}
\end{lstlisting}

Question: \{question\}

Reason \& JSON output:
\end{promptbox}
\caption{The system prompt used for LMM-as-Judge evaluation on 5 dimensions.}
\label{pmt:llm_judge}
\end{figure*}

\begin{figure*}[h]
\footnotesize
\begin{promptbox}{Prompt: Multimodal Adaptation}
**Role:** You are an expert Educational Content Editor specializing in **Multimodal Adaptation**. Your goal is to transform text-only STEM problems into "Visual-Dependency" problems.

**Task:** You will receive a raw text question. You must rewrite it so that the specific **numerical values, geometric properties, or component parameters** are removed from the text and implied to be present in an accompanying image.

**Core Logic:**
The user MUST look at the image to solve the problem. If the text contains all the data, the image becomes redundant. You must fix this redundancy.

---

\#\#\# Step-by-Step Execution Rules

**1. Identification (What to Hide):**
Identify the **"Given Data"** (Explicit Parameters) that can be visualized as labels.

* *Target:* Numbers (e.g., "5 kg", "30°", "10V", "radius 4"), Coordinates (e.g., "(0,2)"), Component types if labeled (e.g., "a 500nm wavelength").

* *Exception:* Do NOT hide the **"Unknown/Target"** (what the user needs to find) or generic constants (e.g., "g = 9.8 m/s²") unless they are specific to the diagram setup.

**2. Removal \& Replacement (How to Hide):**

* **REMOVE** the identified specific values/properties from the text.

* **REPLACE** them with visual pointers such as:
* *"as shown in the figure"*
* *"the labeled angle"*
* *"the indicated dimensions"*
* *"the circuit diagram below"*
* *"shown/depicted"*

**3. Grammatical Repair:**

* Ensure the rewritten sentence flows naturally.
* *Bad:* "A block of [removed] is on a [removed] plane."
* *Good:* "A block of **labeled mass** is on an inclined plane **as shown**."

---

\#\#\# Examples of Adaptation

**Example 1: Geometry**

* **Input:** "In triangle ABC, angle A is 60 degrees and angle B is 45 degrees. Find angle C."
* **Output:** "In the triangle ABC **shown in the figure**, find angle C given **the labeled angles**."
* *(Note: 60 and 45 are hidden).*

**Example 2: Physics (Circuit)**

* **Input:** "A 10V battery is connected to a 5-ohm resistor."
* **Output:** "A battery is connected to a resistor **as shown in the circuit diagram**. Calculate the current."
* *(Note: 10V and 5-ohm are hidden).*

**Example 3: Physics (Mechanics)**

* **Input:** "A ball is thrown with an initial velocity of 20 m/s at an angle of 30 degrees."
* **Output:** "A ball is thrown with the initial velocity and angle **indicated in the diagram**."
* *(Note: 20 m/s and 30 degrees are hidden).*

**Example 4: Math (Simple)**

* **Input:** "Find the area of a circle with radius 4."
* **Output:** "Find the area of the circle **shown below**."
* *(Note: Radius 4 is hidden).*

---

\#\#\# Output Format

Return the result in JSON format:

\begin{lstlisting}[ basicstyle=\linespread{0.75}\ttfamily\scriptsize]
{
  "original_question": "[Input Question]",
  "hidden_parameters": ["List specific values you removed, e.g., '5kg', '30 degrees'"],
  "multimodal_question": "The rewritten text with visual pointers"
}
\end{lstlisting}

---

\#\#\# User Input

**Question:** \{question\}
\end{promptbox}
\caption{The system prompt used for adapting text-only problems into multimodal format.}
\label{pmt:multimodal_adaptation}
\end{figure*}

\clearpage

\end{document}